# High-resolution canopy height map in the Landes forest (France) based on GEDI, Sentinel-1, and Sentinel-2 data with a deep learning approach.

Martin Schwartz[a], Philippe Ciais[a], Catherine Ottlé[a], Aurelien De Truchis[b], Cedric Vega[c], Ibrahim Fayad[d], Martin Brandt[e], Rasmus Fensholt[e], Nicolas Baghdadi[d], François Morneau[f], David Morin[g], Dominique Guyon[h], Sylvia Dayau[h], Jean-Pierre Wigneron[h]

(a) Laboratoire des Sciences du Climat et de l'Environnement, LSCE/IPSL, CEA-CNRS-SCE32UVSQ, Université Paris Saclay, 91191 Gif-sur-Yvette, France.
(b) Kayrros SAS, Paris, 75009, France
(c) IGN, Laboratoire d'Inventaire Forestier, 54000, Nancy, France
(d) INRAE, UMR TETIS, Université de Montpellier, AgroParisTech, CEDEX 5, 34093 Montpellier, France
(e) Department of Geosciences and Natural Resource Management, University of Copenhagen, Copenhagen, Denmark
(f) IGN, Service de l'Information Statistique, Forestière et Environnementale, 45290 Nogent-sur-Vernisson, France
(g) CESBIO, Université de Toulouse, CNES/CNRS/INRAE/IRD/UPS, 18 Av. Edouard Belin, bpi 2801, CEDEX 9, 31401 Toulouse, France
(h) ISPA, UMR 1391, INRA Nouvelle-Aquitaine, Bordeaux Villenave d'Ornon, France

**Abstract**

In intensively managed forests in Europe, where forests are divided into stands of small size and may show heterogeneity within stands, a high spatial resolution (10 - 20 meters) is arguably needed to capture the differences in canopy height. In this work, we developed a deep learning model based on multi-stream remote sensing measurements to create a high-resolution canopy height map over the "*Landes de Gascogne*" forest in France, a large maritime pine plantation of 13,000 km² with flat terrain and intensive management. This area is characterized by even-aged and mono-specific stands, of a typical length of a few hundred meters, harvested every 35 to 50 years. Our deep learning U-Net model uses multi-band images from Sentinel-1 and Sentinel-2 with composite time averages as input to predict tree height derived from GEDI waveforms. The evaluation is performed with external validation data from forest inventory plots and a stereo 3D reconstruction model based on Skysat imagery available at specific locations. We trained seven different U-net models based on a combination of Sentinel-1 and Sentinel-2 bands to evaluate the importance of each instrument in the dominant height retrieval. The model outputs allow us to generate a 10 m resolution canopy height map of the whole "*Landes de Gascogne*" forest area for 2020 with a mean absolute error of 2.02 m on the Test dataset. The best predictions were obtained using all available satellite layers from Sentinel-1 and Sentinel-2 but using only one satellite source also provided good predictions. For all validation datasets in coniferous forests, our model showed better metrics than previous canopy height models available in the same region.

**Keywords: Forest height, GEDI, Sentinel-1, Sentinel-2, U-Net, Deep Learning, Landes forest, Forest Inventory, 3D Stereo**

# 1 Introduction

Forest biomass plays a crucial role in the global carbon cycle, and its conservation or increase is an essential element of land-based mitigation policies (Griscom et al., 2017; IPCC, 2019; Pan et al., 2011; UNFCCC, 2015). In intensively managed forests of Europe, where forests are divided into stands of small size, a relatively high spatial resolution (10 - 20 meters) is needed to capture the difference in biomass between adjacent stands or within stands in case of heterogeneous forest structure. Top canopy height, associated with other forest parameters such as tree species, can be a relatively accurate proxy for forest biomass estimation (Duncanson et al., 2022). Forest inventories have been the only method to estimate forest biomass and height in the past. They provide reliable statistical information on forests over large regions but are not designed to produce high-resolution maps. For a few decades, remote sensing data opened the possibility of height estimation at a finer scale with more collected data. Airborne LiDAR (ALS) can scan a forest and provide accurate height estimations consistent with forest inventory measurements at a high spatial resolution. These data, used as reference height for models using space-borne images, have a high potential to accurately describe forest structures (Wilkes et al., 2015). However, their acquisition is costly, and measurement campaigns are sparse in time, which does not enable up-to-date maps. On the other hand, space-borne measurements have a lower spatial resolution but a higher temporal resolution and global coverage. They have the potential to map forest properties at a global scale monthly or yearly like the product from (Hansen et al., 2013), which derived global forest annual loss and probability of gain at 30 m resolution from 2000 to 2012 from Landsat imagery. Additionally, spaceborne LiDARs such as ICESat demonstrated the potential to map forest height (Copernicus Land Monitoring Service, 2018). Global and local maps of forest biomass and height have been developed based on various remote sensing approaches, often with worldwide coverage and a medium to low resolution (~100 to 1000 m). They are often based on spaceborne or airborne observations, calibrated with in-situ measurements.

The GEDI (Global Ecosystem Dynamics Investigation) LiDAR mission, developed and operated by NASA onboard the International Space Station (ISS) since 2019, has produced accurate point-wise observations of forest structure (Dubayah et al., 2020). Associated with other space-borne and airborne data, this instrument has shown promising capabilities for height mapping (Fayad et al., 2021a; Lang et al., 2022b, 2022a; Potapov et al., 2021). Sentinel-1 (S1) and Sentinel-2 (S2) are two satellite missions of ESA's Copernicus program for earth observation. They provide measurements of earth's radar (Synthetic Aperture Radar) backscattering coefficients (Sentinel-1) or multi-spectral reflectance (Sentinel-2) at 10 m resolution with a revisit interval of approximately five days since 2015 and have already been used to map biomass and height at high resolution (Lang et al., 2019; Li et al., 2020; Morin et al., 2019). Combinations of GEDI and Sentinel-2 have already been proposed for mapping crop height (Tommaso et al., 2021) and forest canopy height (Lang et al., 2022a), and its potential to estimate forest height was evaluated by Pereira-Pires et al. (2021) with linear and exponential regressions.

Machine learning has proven some solid results for forest parameter estimations in previous remote sensing studies (Fayad et al., 2014; Morin et al., 2019; Potapov et al., 2021). More recently, deep learning and, more specifically, convolutional networks (CNN) have provided a new set of tools allowing remote sensing research to process large amounts of training data for more accurate predictions (Ball et al., 2017; Zhu et al., 2017). CNNs (LeCun et al., 2015) have significantly increased accuracy related to image interpretation tasks. Thanks to a series of linear operations (convolutions) and non-linear "activation" functions, these models can learn multi-scale image features, such as image texture, that are then used to carry out predictions. CNNs are already widely used for object detection or scene classification tasks and have proven their efficiency in remote sensing (Zhu et al., 2017). However, few studies have used these models for regression tasks like tree height mapping (Dalagnol et al., 2022; Illarionova et al., 2022) and, to our knowledge, no studies have used simultaneously GEDI, Sentinel-1, Sentinel-2 and a CNN model to estimate canopy height up to now.

Here, we introduce a new methodology that leverages the potential of GEDI to be used as reference height to train a deep-learning model. The methods combine GEDI's height pointwise measurements and S1 and S2 images at a high temporal and spatial resolution to create wall-to-wall height maps at 10 m scale, based only on space-borne data, over a large forest area in France. The region analyzed is the largest western European plantation forest: the *Landes de Gascogne* (referred to as Landes forest in the following), a maritime pine plantation located in the South-West of France. Our model is trained on seven combinations of S1 and S2 layers. The retrieval results are evaluated using several in-situ datasets: a dense maritime pine inventory performed in 2016 (GLORIE), the French National Forest Inventory (NFI) distributed all over the study area, and height map based on 3D stereo height reconstruction from Skysat imagery at a specific location within the Landes forest. Additionally, the model is compared to three canopy height maps available in the area of interest.

# 2 Materials & Methods

This study relies on the use of canopy height measured by the GEDI sensor, a space-borne LiDAR onboard the ISS (Dubayah et al., 2020). We used GEDI's $RH_{95}$ variable defined in 2.2.1 as reference height samples and Sentinel-1 and Sentinel-2 images (2.2.2 and 2.2.3) as predictors for a deep learning U-Net framework (2.4.1) to provide a gridded map at 10 m resolution of the *Landes de Gascogne* area. Fig. 1 describes this general workflow.

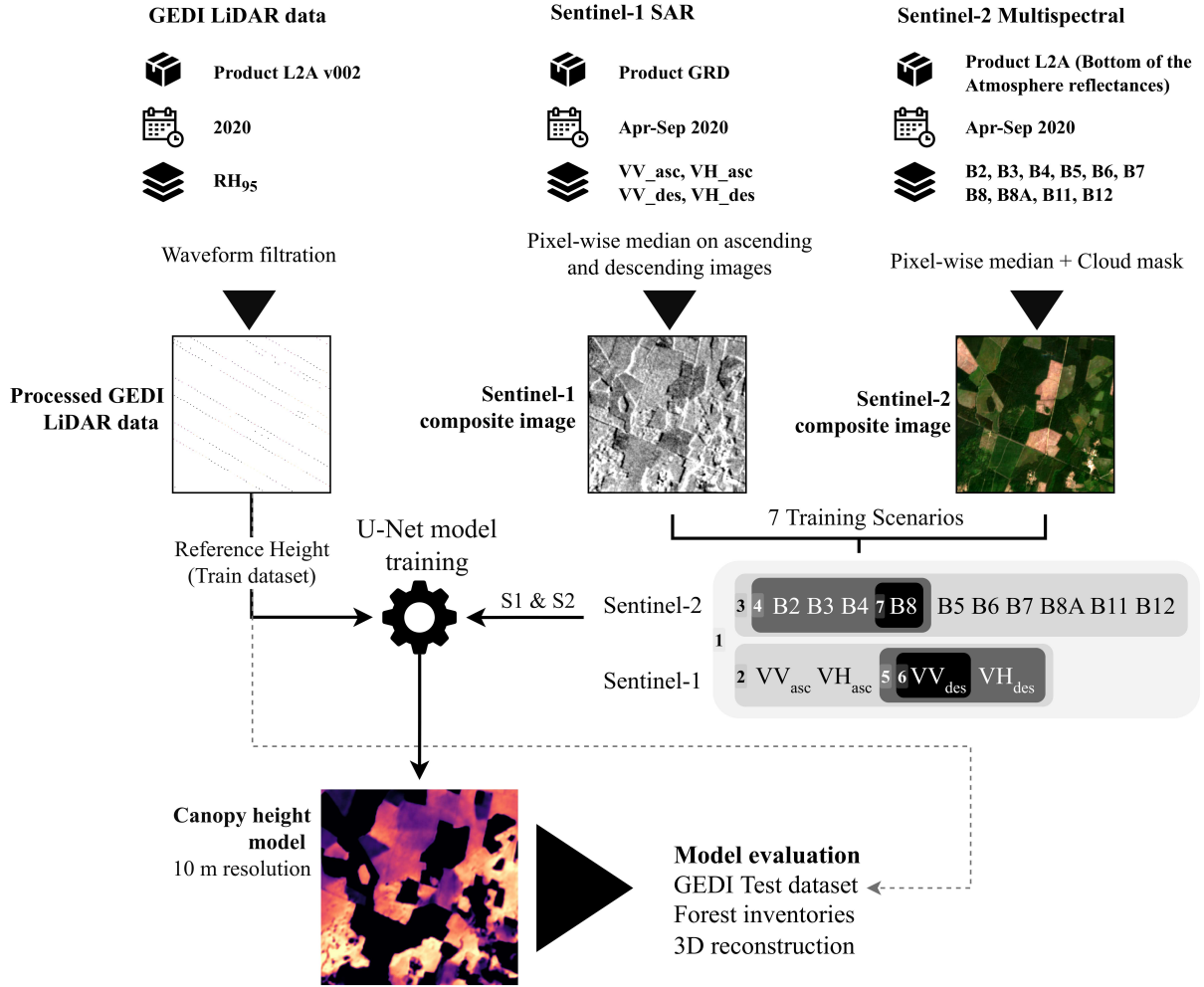

*Figure 1. General workflow showing the data preprocessing steps, the U-Net model training, and the evaluation strategy. More details for the U-Net training process are presented in Fig. 5. The seven training scenarios correspond to the layers from S1 and S2 used in the training process: 1 - all layers; 2 - all S1; 3 - all S2; 4 - S2 RBG + NIR, 5 - S1 VV_des + VH _des; 6 - VV_des; 7 - S2 NIR.*

## 2.1 Study area

The Landes forest is located in a flat region with an oceanic climate and sandy soils in the South-West of France (Fig. 2). It is the largest European plantation (~ 1 million hectares) composed of 90% of maritime pine (*Pinus Pinaster*). The remaining forested part consists of broadleaved forests, including several species of oaks, mainly located around rivers. This forest is intensively managed: thinning occurs every 5-10 years, clear-cuts are performed after 35-50 years, and they are replanted within 2-3 years. Forest management leads to very homogeneous tree repartition within stands but also to a high heterogeneity between separate forest parcels. Understory vegetation comprises various woody shrubs and perennial herbaceous plants (fern). It is regularly cleared during the first years following reforestation.

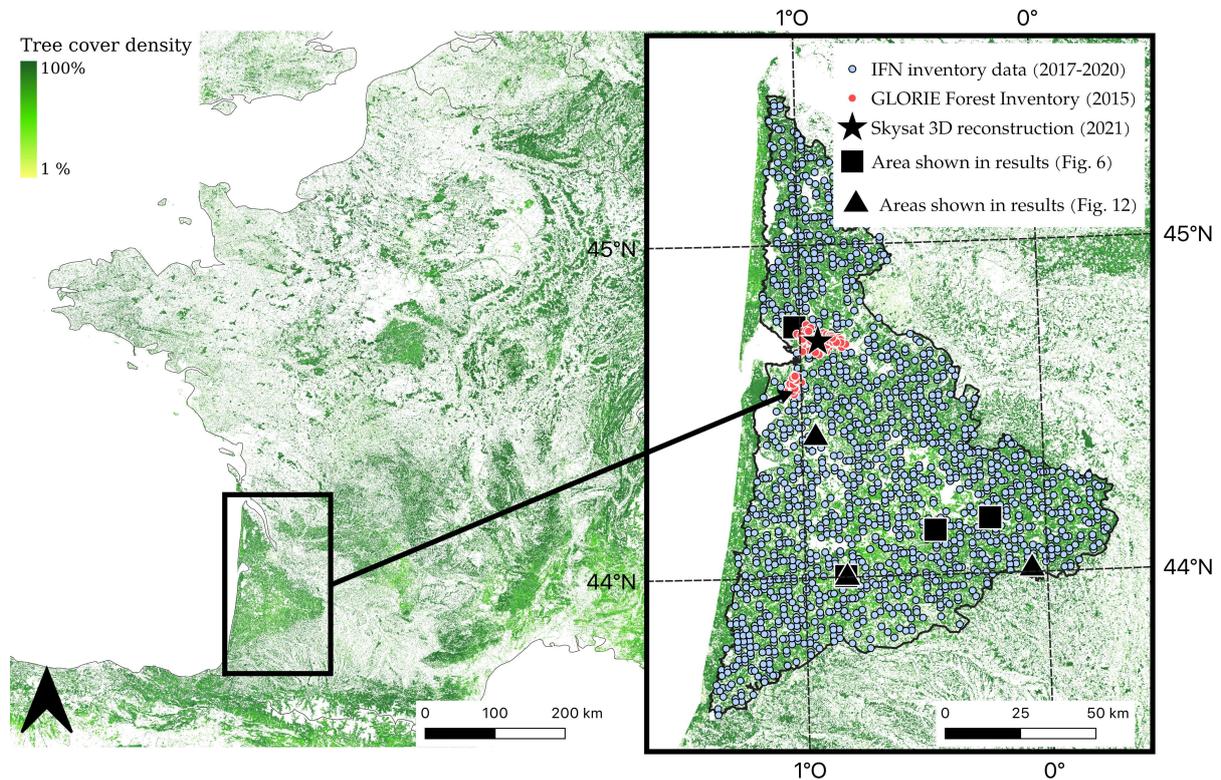

*Figure 2. Study site, known as "Les Landes de Gascogne" (referred to as Landes forest). Forest inventory plots, GLORIE inventory, Skysat 3D reconstruction, and forest areas considered in the results are shown. Greenness indicates tree cover density (Copernicus Land Monitoring Service, 2018)*

## 2.2 Datasets used to train the model

### 2.2.1 GEDI

Data from the Global Ecosystem Dynamics Investigation (GEDI) are used as a reference variable to model a continuous height map at 10 m resolution. GEDI is operated by NASA and produces high-resolution LiDAR observations of the Earth's vertical structure (Dubayah et al., 2020). This spaceborne infrared LiDAR, deployed in December 2018 on the ISS, provides energy return waveforms (L1B product) and derived metrics such as canopy relative height (RH) and plant area index (L2A/L2B product) that describe the vertical forest structure within 25 m diameter circular footprints. The instrument acquires data over eight tracks with a footprint spacing of 60 m along the track and 600 m across tracks. Due to the ISS orbit, it covers latitudes between 51.6° South and 51.6° North. The precision of the horizontal footprint location, initially around 20 m, has been improved in GEDI's second release to a value of 10 m, as shown in Appendix 1 (Dubayah et al., 2021). The GEDI L2A product provides RH metrics representing the height relative to the ground of n% ($RH_n$ with n in the 0 to 100 range) of the total returned energy between the top of the canopy and the signal end. These metrics are extracted from the raw waveforms with six different algorithms representing different combinations of thresholds and smoothing settings. They can vary from one algorithm to another depending on the forest type (Adam et al., 2020). In our study area, most footprints had the same RH values for most algorithms. Hence, we chose to use the algorithm selected by NASA in the "selected_algorithm" variable in GEDI data, which corresponds mainly to algorithm 1 over our study area. In theory, $RH_{100}$ should represent the top of canopy height. Still, this metric is affected by noise from atmospheric disturbances, uncertainties on the position of the detected ground return, vegetation, and ground variability. Here, we used $RH_{95}$, as it has showed a better correlation with other height sources and has been used as a proxy for height in previous studies (Fayad et al., 2021b; Potapov et al., 2021) even though other similar metrics such as $RH_{98}$ have also been used (Lang et al., 2022b). Due to the LiDAR properties, this measure is intended for vegetation and may represent confusing results for bare soil or water bodies (Beck et al., 2020). Indeed, these surfaces mirror the transmitted waveforms that have a pulse width of ~ 15 ns which corresponds to a ~ 2.25 m wide waveform (Dubayah et al., 2020). In total, 526,449 footprints from the GEDIv002 L2A product (Dubayah et al., 2021) were downloaded from NASA's EarthDataSearch website (https://search.earthdata.nasa.gov/search) for this study, covering the entire area of interest for 2020. Due to atmospheric perturbations, some waveforms could not be used to give information on the vertical forest structure. Therefore, several filtering criteria were applied to remove unusable waveforms: (1) When the *quality_flag* provided in the GEDI data was set to zero. (2) When one of *toploc, botloc, num_detectedmodes, $RH_{100}$* provided metrics had a null value. (3) When the ratio defined as the maximum amplitude of the waveform divided by the standard deviation of noise was lower than 30 (arbitrary value that removed 3% of the remaining waveforms). As the study area is mostly flat, terrain was not considered in the data filtering. Moreover, recent studies (Fayad et al. 2021c) showed that GEDI canopy height retrievals were not affected by small slopes (<45%). After these operations, 175,511 GEDI valid waveforms (33%) were kept in 2020. These GEDI footprints were then spatially separated into Train, Validation, and Test datasets. For this, the study site was divided into 117 tiles of 100 km² and randomly separated into 91 Train tiles, 15 Validation tiles, and 11 Test tiles so that it corresponds respectively to 75%, 15%, and 10% of the GEDI footprints (Fig. 3a). The temporal distribution of the data is uneven and some months (July, ~ 16% of the data) are more represented than others (January, ~ 4% of the data). The spatial distribution of the data is also uneven due to the ISS trajectory and the northern tiles of the study area contain more footprints than other tiles. The spatial and temporal distributions of these data are shown in Appendix 5. To evaluate our model on forested areas only, in Test and Validation tiles ,we removed the GEDI footprints located in pixels where the tree cover has a null value in the Copernicus tree cover density map (Copernicus Land Monitoring Service, 2018). For the Train dataset, we kept all data, thus leading to more realistic results over non-forest areas. The height repartition of data in Train, Validation, and Test datasets is presented in Fig.3b.

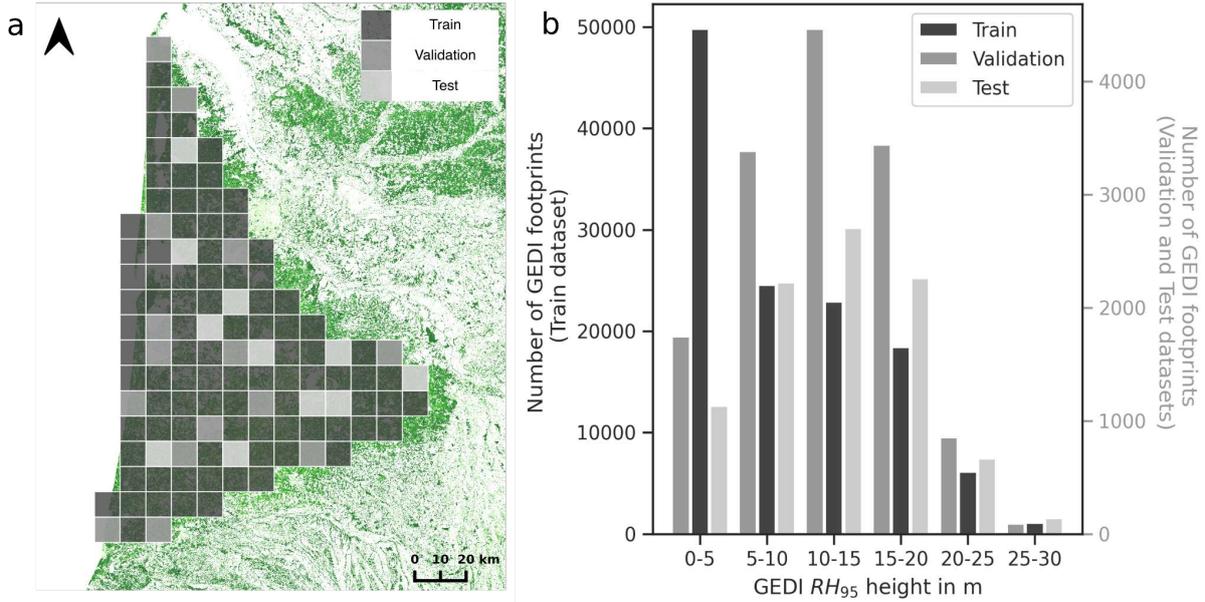

*Figure 3. (a) Repartition of the area of interest into 100 km² tiles divided into Train, Validation, and Test datasets. (b) Count of GEDI waveforms per height in Train, Validation, and Test datasets. Gedi footprints from non-forest areas were removed from Validation and Test datasets.*

### 2.2.2 Sentinel-1

Sentinel-1 (S1) is a C-band Synthetic Aperture Radar (SAR) mission composed of 2 satellites with a sun-synchronous orbit and 12 days repeat cycle (Sentinel-1A, Sentinel-1B) launched respectively in 2014 and 2016 by ESA. Here, we used the Ground Range Detected (GRD) scenes with dual-band cross-polarization (Vertical-Vertical + Vertical-Horizontal bands at 10 m resolution). They were preprocessed with the Sentinel-1 toolbox in Google Earth Engine, which includes thermal noise removal, radiometric calibration, and terrain orthorectification as specified at https://developers.google.com/earth-engine/guides/sentinel1. The S1 data set provides backscattering coefficients (dB) which are a measure of the backscattered microwave radiations (backward direction) emitted by the radar system.

Using the Google Earth Engine (GEE) online platform, we selected 280 images covering the entire area of interest over a five-month period in 2020 (2020-05-01 to 2020-10-01). This period was selected to have enough data to reduce the speckle observed in raw S1 images but still not too large to keep a similar aspect of the forest canopy (leaf-on season) to calculate a median composite. We then separated them between ascending and descending orbits (140 images per category) and calculated the per-pixel median of these image time series, thus creating a single composite image of 4 layers: VV_ascending, VH_ascending, VV_descending, and VH_descending at 10 m resolution. Each pixel of the composite image corresponds to the median value of pixels from ~25 images at different dates. The median is little influenced by extreme values and reduces the potential soil and vegetation moisture-related effect that could be observed on raw S1 images. These bands were then restrained to values between -30 dB and 0 dB and scaled to a 0 to 1 interval in order to have a similar range of values as the Sentinel-2 input data (See 2.2.3).

### 2.2.3 Sentinel-2

The Sentinel-2 (S2) mission is part of the European Spatial Agency's (ESA) Copernicus program for Earth Observation (EO). It is composed of two satellites, Sentinel-2A, launched in 2015, and Sentinel-2B, launched in 2017, on a sun-synchronous orbit. S2 provides multi-spectral images of the Earth surface reflectance with a low revisit interval of ~ 5 days, including images from both satellites, thanks to a large swath of 290 km. The L2A product used here provides bottom of the atmosphere reflectance, processed from L1C (Top of the atmosphere reflectance) with ESA's Sen2Cor processor (Main-Knorn et al., 2017). It comprises 13 spectral bands ranging from 10 m to 60 m resolution in visible, near-infrared (NIR), and short-wave infrared (SWIR).

Based on the Google Earth Engine (GEE) online platform, we selected 172 images within the entire area of interest during the same period as the S1 images (2020-05-01 to 2020-10-01) with less than 50% of clouds. After applying a cloud mask, we created a single composite image by taking the per-pixel median value over the image time series . We chose the median composite to be less sensitive to outlier pixels such as pixels affected by clouds or cloud shadows that could remain even after the cloud mask. The median value computed for each pixel comes from 16 to 72 images depending on the pixel location because of differences in image overlap. The ten following spectral bands (10 and 20 m resolution) were kept and resampled to 10 m if necessary: B2: Blue, B3: Green, B4: Red, B5-B6-B7: Red edge, B8: Near Infrared (NIR), B8A: "narrow" NIR, B11-B12: Short Wave Infrared (SWIR). We did not calculate any hand-crafted features such as vegetation indices or texture features insofar as we used a deep learning network (See 2.4.1) which calculates the best-suited features by itself thanks to convolutional filters. The input data of a neural network should be on a similar scale to help stabilize the gradient descent step in the training phase. Thus, it is important to have the same range of values for the different inputs of our model (S1 and S2). Sentinel-2 reflectance values in GEE are given in digital units (DN) between 0 and 10000 (https://developers.google.com/earth-engine/datasets/catalog/COPERNICUS_S2_SR) where DN = 10000 * Reflectance.  Forest reflectance typically ranges from 0 to 0.3 (DN from 0 to 3000) depending on the spectral bands. To have a better contrast and a more suited distribution for deep learning without a long tail, we clipped the DN values in the 0~5000 interval (values higher than 5000 were set to 5000). Then we divided all the values by 5000, creating a range of values from 0 to 1. Thanks to filtering, cloud mask, and median average, the resulting composite image was exempt from cloudy areas.

## 2.3 Evaluation Datasets

To compare our height predictions with independent sources, we used three complementary datasets covering different spatial scales: a very high-resolution 3D stereo reconstruction from Skysat imagery, a dense local forest inventory (GLORIE) performed near the Bordeaux area, and the French national forest inventory data from IGN (Institut national de l'information géographique et forestière) spanning the entire area of interest (Fig. 2). Additionally, we compared our predictions to canopy height maps computed from previous local and global studies using other techniques or data sources (Lang et al., 2022a; Morin et al., 2019; Potapov et al., 2021)

### 2.3.1 Forest inventory data

The French national forest inventory (NFI), performed yearly by the French Geographical Institute (IGN, Institut Géographique National), carries out forest measurement campaigns in France (IGN, 2022). Every year, 30 m diameter circular plots are sampled in one-tenth of a grid that covers the national territory. These forest plots are distributed in the Train, Validation, and Test tiles (see 2.2.1). The dominant height variable provided by IGN for these forest plots corresponds to the estimated mean height of the 100 highest trees within a surface area of 1 hectare. In order to avoid a too long time difference between the date of inventory and the date of our prediction, we selected only plots that were measured between 2017 and 2021. To compare these heights to our predicted heights, we used the mean value of our 10 m x 10 m pixels that intersect the 30 m circular plots of the IFN, which corresponds to the mean of ~ 9 pixels. We also tested other comparison methods, e.g., taking the maximum value of these pixels and taking the single 10 m x 10 m pixel value at the center of the plot. These different comparison methods yielded very similar results and we used the mean value for this study.

The GLORIE local forest inventory was acquired in winter 2015-2016 before the beginning of tree growth (Motte et al., 2016; Zribi et al., 2019). It includes measurements of, among others, tree heights, DBH, and tree density. The 99 forest plots (location indicated in Fig. 2) were selected from forest stands of maritime pines, trying to cover all the range of forest structures in this region. The GLORIE forest plots are located within the Train tiles (see 2.2.1). The dominant height variable provided in this dataset was calculated as the mean height of the two largest trees (higher diameter at 1.30 m) among the ten trees closer to the center of the forest plot. Here, we computed the mean value of the height pixels predicted by our model within a 25 m diameter circle (typical size of these forest plots) around the location of the forest inventory plots and compared it to the dominant height measured in the inventories.

Although forest inventory canopy height data were considered as ground truth to evaluate our model, uncertainties are associated with them. Forest inventory accuracy on height measurements strongly relies on the surveyors and the method used (Berger et al., 2014; Jurjević et al., 2020; Kitahara et al., 2010). Internal IGN studies based on forest inventory control, re-measuring ~ 4000 trees each year, have shown that the typical standard error on tree height estimation ranges from 1.4 m to 1.75 m.

### 2.3.2 Stereo 3D reconstruction from Skysat imagery

Skysat is a constellation of optical sub-meter resolution EO satellites owned by Planet Labs, providing high-resolution imagery in panchromatic, visible, and infrared bands at 0.9 m resolution. Its geolocation accuracy is 3.4 m for all images (Saunier et al., 2021), but it should be better on the cloud-free images used here. The stereo 3D height reconstruction technique of de Franchis et al. (2014) uses multiple image acquisitions with different view angles to reconstruct 3D objects with a dispersion error of ~ 0.5 m. Here, we applied this technique to reconstruct canopy height in a small area of 11 km² (Location indicated in Fig. 2) and compared it to our model outputs. We first created a digital surface model (DSM) at 0.8 m resolution from point-cloud 3D reconstruction based on Skysat images acquired in 2021. Then we used a cloth simulation algorithm (Zhang et al., 2016)) to select ground points from the 3D point cloud. Finally, we used a Laplace interpolation to create a terrain model from those points that we subtracted from the raw DSM to obtain a canopy height model (CHM). To compare it to our model, we resampled this CHM to a 10 m grid aligned with our prediction map. For this, we took the maximum value of the CHM within each 10m x 10m grid cell, which corresponds to the top canopy height and is thus relevant for a comparison with our $RH_{95}$-based model. As forest stands in the Landes region are very homogeneous, we carried out two types of comparison with our model: (1) a simple per-pixel comparison of the two maps. (2) A comparison at the forest stand level by taking the median value over forest stands that we delimited manually (Fig. 4).

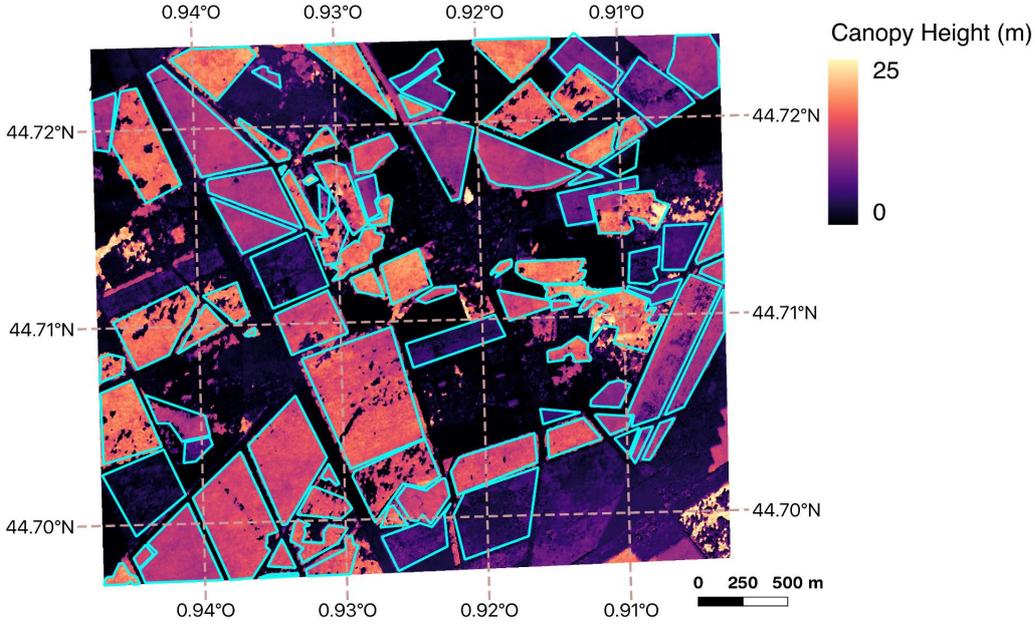

*Figure 4. Stereo 3D reconstruction from Skysat imagery (0.8 m resolution). Forest stands estimated from manual delimitation are represented by blue polygons.*

### 2.3.3 Canopy height maps from previous studies

We compared our model predictions to three canopy height maps from independent studies available in the Landes forest (Lang et al., 2022a; Morin et al., 2019; Potapov et al., 2021). Maps from Potapov et al. (2021) and Lang et al., (2022a) are two global canopy height maps, hence non-specific to the study area, based on GEDI and optical data only. Potapov et al. (2021) have mapped forest height globally in 2019 at 30 m resolution, using Landsat-8 images to extrapolate canopy height from GEDI footprints with a bagged regression trees ensemble method. Lang et al. (2022a) have used Sentinel-2 images to extrapolate canopy height from GEDI footprints with a deep fully convolutional network and create a global canopy height map for 2020 at 10m resolution. The map from Morin et al. (2019) is a local canopy height product specific to maritime pine homogeneous plantations in the Landes forest. Contrary to the two previous maps that used GEDI, reference height measurements were based on a small local forest inventory of maritime pine only (GLORIE inventory, see 2.3.1). These forest inventory measurements were used with remote sensing data (Sentinel-2, Sentinel-1, ALOS-PALSAR…) and a machine learning model (Support vector machine) to produce a canopy height map of the Landes forest in 2016, only valid for maritime pine plantations.

## 2.4 Methodology

### 2.4.1 U-Net model description

In this study, we performed a pixel-wise regression, which is the process of attributing a particular value (tree height) to each pixel instead of attributing a label to the whole image. This task can be addressed by fully convolutional networks (FCNs) (Long et al., 2015) that have been adapted from classical CNNs. Here, we use a U-Net model adapted from Milesi (2022) which is a FCN that outperformed previous models in speed and accuracy and requires fewer training examples thanks to its U-shape architecture (Ronneberger et al., 2015). This model consists of a contracting path (left) and an expansive path (right) which gives it its "U" shape and enables the model to extract relevant information at different spatial scales. In the contracting path, the input image goes through two 3x3 convolutions followed by a rectified linear unit (ReLU) and is then downsampled with a 2x2 max pooling operation with stride 2. This is repeated four times, and at each step, the number of feature channels is multiplied by two. In the expansive path, the image first goes through a bilinear upsampling, then it is concatenated with the corresponding image of the contracting path and finally goes through two 3x3 convolutions followed by a ReLU like in the contracting path. This step is also repeated four times and at each step, the number of feature channels is divided by four instead of two because of the concatenated image. The final layer consists of a 1x1 convolution that creates an output image with the same height and width as the input image. After the training process, this output will represent the canopy height map derived from the input image. In total the network has 18 convolutional layers and ~17 Million trainable weights. The exact architecture of the prediction model we developed can be found in Appendix 2 (it will be referred to as the "FCN model" in the following).

### 2.4.2 Training process

The objective of the training process was to adjust all the trainable weights of the FCN model so that when the FCN model takes a multi-channel image composed of S1 and S2 layers as input, it outputs the corresponding canopy height map. One single FCN model was trained on all the Train tiles. To achieve this, we followed the process described in Fig.5: (1) We randomly selected one of the 91 Train tiles. This random selection was weighted by the number of GEDI footprints in each tile. (2) We took a random 2560x2560 m subset from this tile with at least one GEDI footprint inside. This technique increases the number of different images used as input of the model and contributes to reduce model overfitting (3) The corresponding 256x256 pixels image composed of S1 and S2 layers was used as input of the U-Net. In the case where we chose to use all S1 and S2 bands, this image was composed of 14 layers. (4) The GEDI $RH_{95}$ values in this 2560x2560 m subset were rasterized on a 10 m grid aligned with S1 and S2. To do so we used the *Rasterio.features.rasterize* function from the Rasterio library in python. For each footprint, the $RH_{95}$ value was rasterized on the pixel where the center of the circular footprint was located. As the GEDI footprints correspond to 25 m diameter circles with a 10 m uncertainty on their location, we acknowledge that our rasterization process at 10 m could create label noise, especially in forests with a very heterogeneous canopy. Still, training at 10 m was shown to give better results than training at 20 m (see Appendix 7). (5) The FCN model output was compared to the reference height from the rasterized GEDI data with a mean absolute error (MAE) loss only for pixels where a $RH_{95}$ height value was available. (6) The gradient of this loss was calculated with respect to each of the model weights and weights were then adjusted accordingly. This process is called "loss backpropagation" and is a key element in the training of neural networks. Here it was performed with a SDG optimizer (momentum of 0.9) with a cyclic learning rate scheduler "triangular2" (Smith, 2017). The base value was set to $1.10^{-7}$, the max value to 0.1, with 320 steps for a half cycle. After each epoch of 960 images (32 batches of 30 images), we calculated the MAE loss on all the validation tiles. After ~100 epochs, which corresponded to ~2 hours, this validation loss was stabilized and we manually stopped the training process. We also tried other types of learning rate schedulers but this one led to faster convergence and similar results. The training process was done with Amazon AWS cloud platform on a GPU NVIDIA Tesla T4 (16 GB). We used the Pytorch library, an open-source machine learning framework in Python, to implement the FCN model.

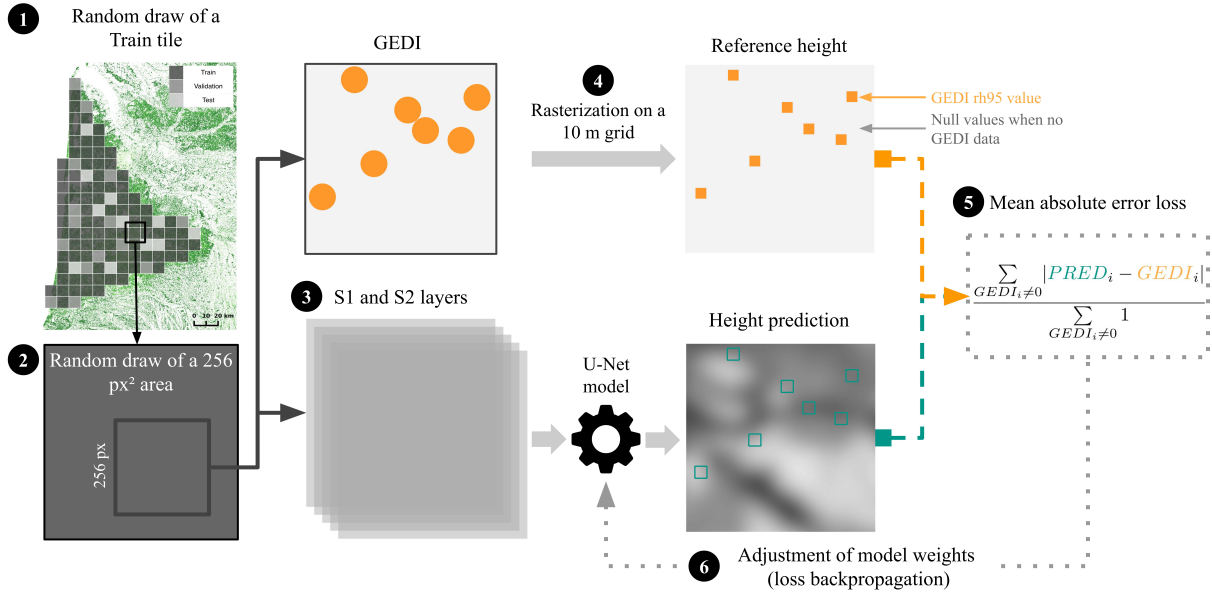

*Figure 5. Computation of the mean absolute error (MAE) loss in the FCN prediction model training process. The numbers in the black circles correspond to the steps described in section 2.4.2 (Training process).*

### 2.4.3 Training scenarios

To understand the importance of each spectral (Sentinel-2) and polarization (Sentinel-1) information, we designed seven training scenarios based on combinations of different spectral and polarization layers. The first FCN model was trained on all layers (10 from S2 and 4 from S1). The second and third FCN models were based on one source of data only (either S2 or S1). The other FCN models were trained on subsets of S2 (only 10m resolution bands or only B8 (NIR band)) and S1 bands (only descending orbit, only VV polarization for descending orbit) respectively (See Fig. 1 for a summary of the 7 scenarios used).

### 2.4.4 Used metrics

To evaluate our FCN model against the validation datasets, we used several metrics, namely the Mean Absolute Error (MAE), the Root Mean Squared Error (RMSE), the Mean Error (ME) that indicates the bias, and the determination coefficient ($R^2$) that is a measure of the correlation between predicted height and true height values. Additionally, in order to have more information on the source of errors, we decomposed the Mean Squared Error (MSE) into three additive terms: Squared Bias (SB), Squared Difference between Standard Deviations (SDSD), and Lack of Correlation weighted by the Standard deviations (LCS) as proposed by (Kobayashi and Salam, 2000) where MSD = SB + SDSD + LCS. SB indicates the bias, SDSD shows how the FCN model is able to simulate the magnitude of the fluctuation between the n measurements and LCS is the ability of the FCN model to simulate the fluctuations across the n measurements. The detailed formulas for these metrics are provided in Appendix 3.

# 3 Results

## 3.1 Canopy height model

In a first step, we present the results for Scenario 1 where the FCN model was trained using all Sentinel-1 and Sentinel-2 layers (as shown later in a detailed comparison (Section 3.4) between the seven scenarios, Scenario 1 provided best prediction results). Fig. 6 illustrates four examples of predictions for different types of forested areas, while Fig. 7 shows our prediction map for the whole Landes forest in 2020.

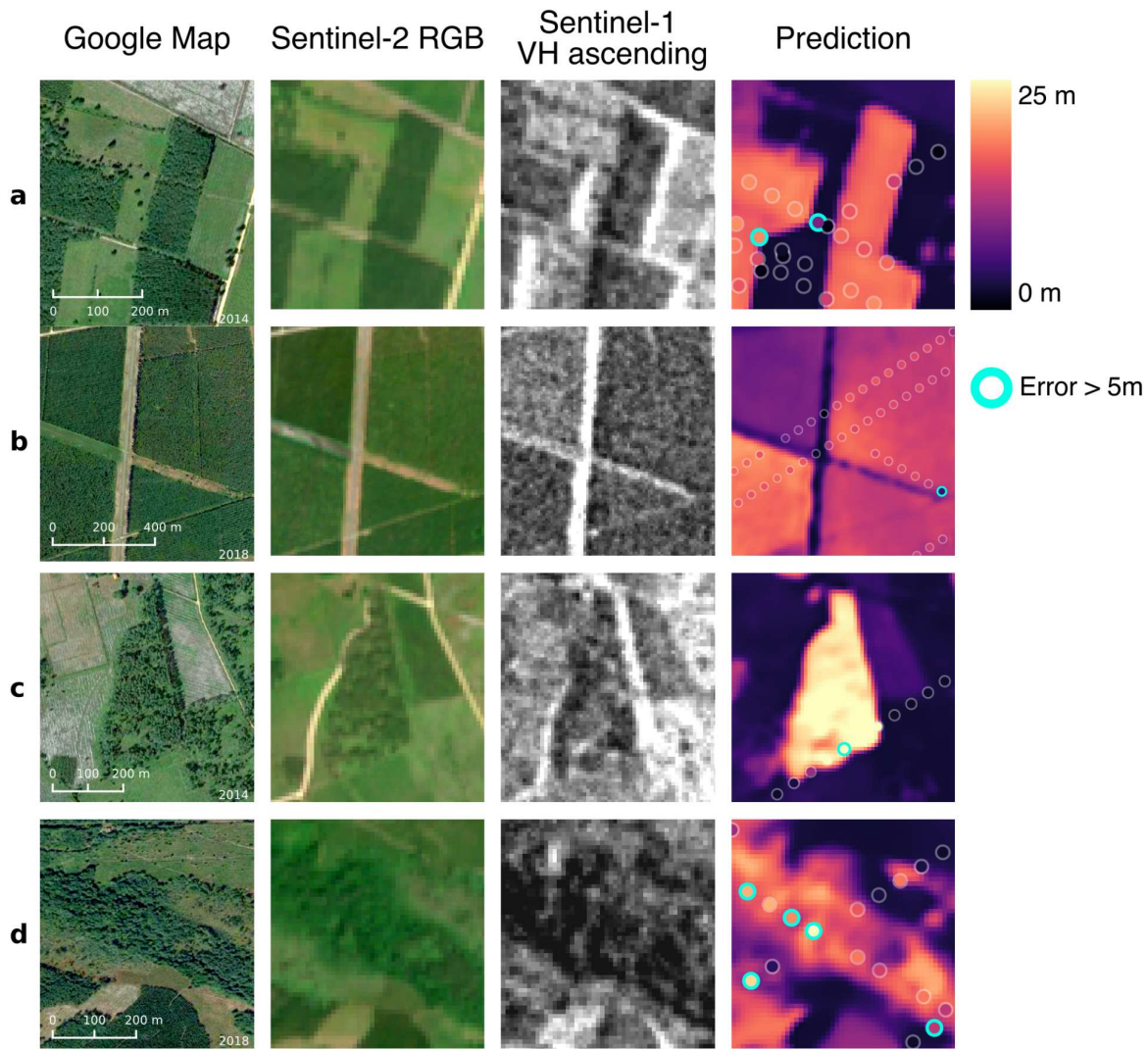

*Figure 6. Model input data and model predictions on four different areas in the Test tiles. The first three columns correspond to Google Map, Sentinel-2 RGB, and Sentinel-1 VH ascending images and the last column corresponds to our predicted map of forest height where GEDI height values ($RH_{95}$) can be identified by circles. When the predicted and the GEDI height are different by at least 5 meters, the GEDI footprint is circled in blue. (a) and (b): maritime pine forest stands of different heights. (c) old forest stands of maritime pine. (d) deciduous forest.*

Overall, even though our FCN prediction model was trained on sparse reference data (eg. rasterized GEDI $RH_{95}$ values in the Train tiles), it is able to predict a continuous canopy height map, where forest structures and other landscape features are recognizable and look similar to the image input. Fig. 6a shows that the FCN model is able to retrieve high height differences between forest from non-forest areas. When compared to the GEDI Test dataset for maritime pine plantations (Fig. 6a, b, c), most errors are below 5 m except in areas close to forest borders. The within-stand homogeneity and across-stand heterogeneity are well captured by the FCN model over coniferous forest stands with different ages and heights (Fig.6b). Fig. 6d shows an example obtained in an area including deciduous trees along a river path. We can see a higher number of differences higher than 5 m when comparing to the GEDI Test footprints and the predicted forest delimitations are less precise.

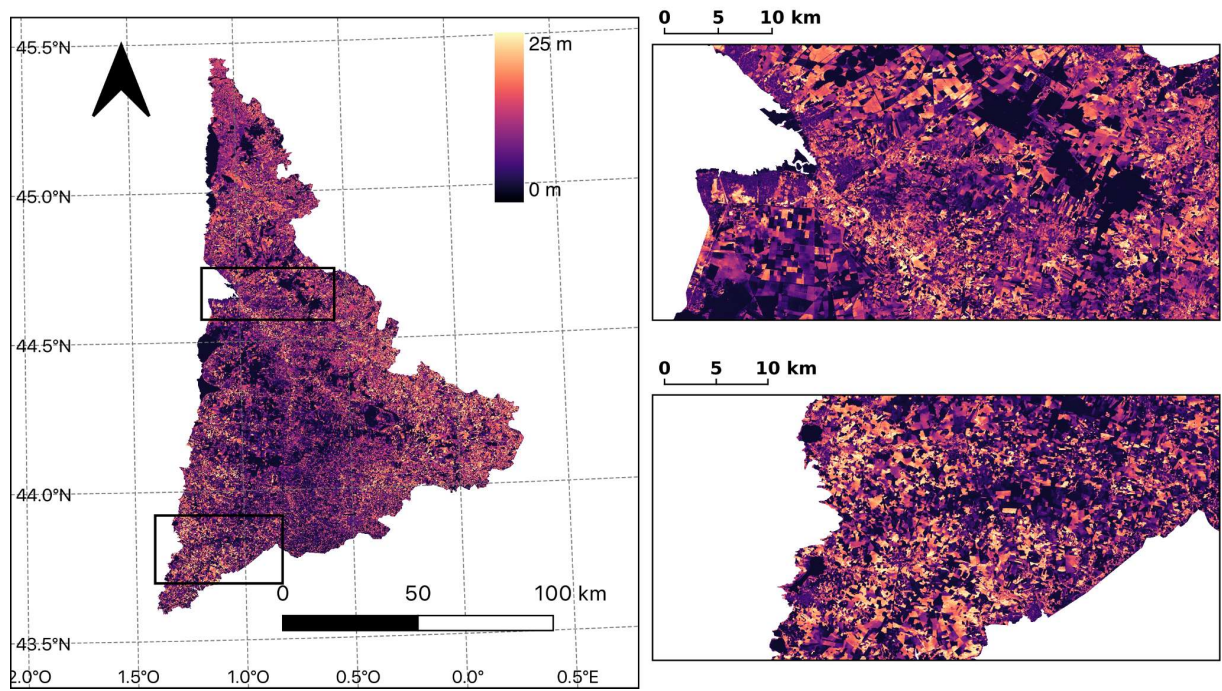

*Figure 7. Map of forest predicted canopy height over the Landes forest for 2020 by the FCN model for the Training Scenario 1: 10 layers from Sentinel-2 + 4 layers from Sentinel-1.*

## 3.2 GEDI Test set evaluation

We compared the predictions of our FCN model (Scenario 1) to the $RH_{95}$ values for the Test dataset. We obtained a Mean Absolute Error (MAE) of 2.02 m, RMSE = 2.98 m, and $R^2$ = 0.73 (Fig. 8). The FCN model has a slight tendency to underestimate higher heights (ME = - 2.5 m for trees between 20m and 25 m) and predictions for lower heights start around 2.5 m due to the $RH_{95}$ properties (see 2.2.1). Predictions are relatively well scattered around the y=x axis with the exception of a group of points located on the left part of the graph where the predictions overestimated the reference height values. Boxplots in Fig.8a reveal almost no bias for height values between 5 m and 20 m (ME = - 0.36 m, MAE = 1.80 m) which accounts for 79% of the GEDI footprints.

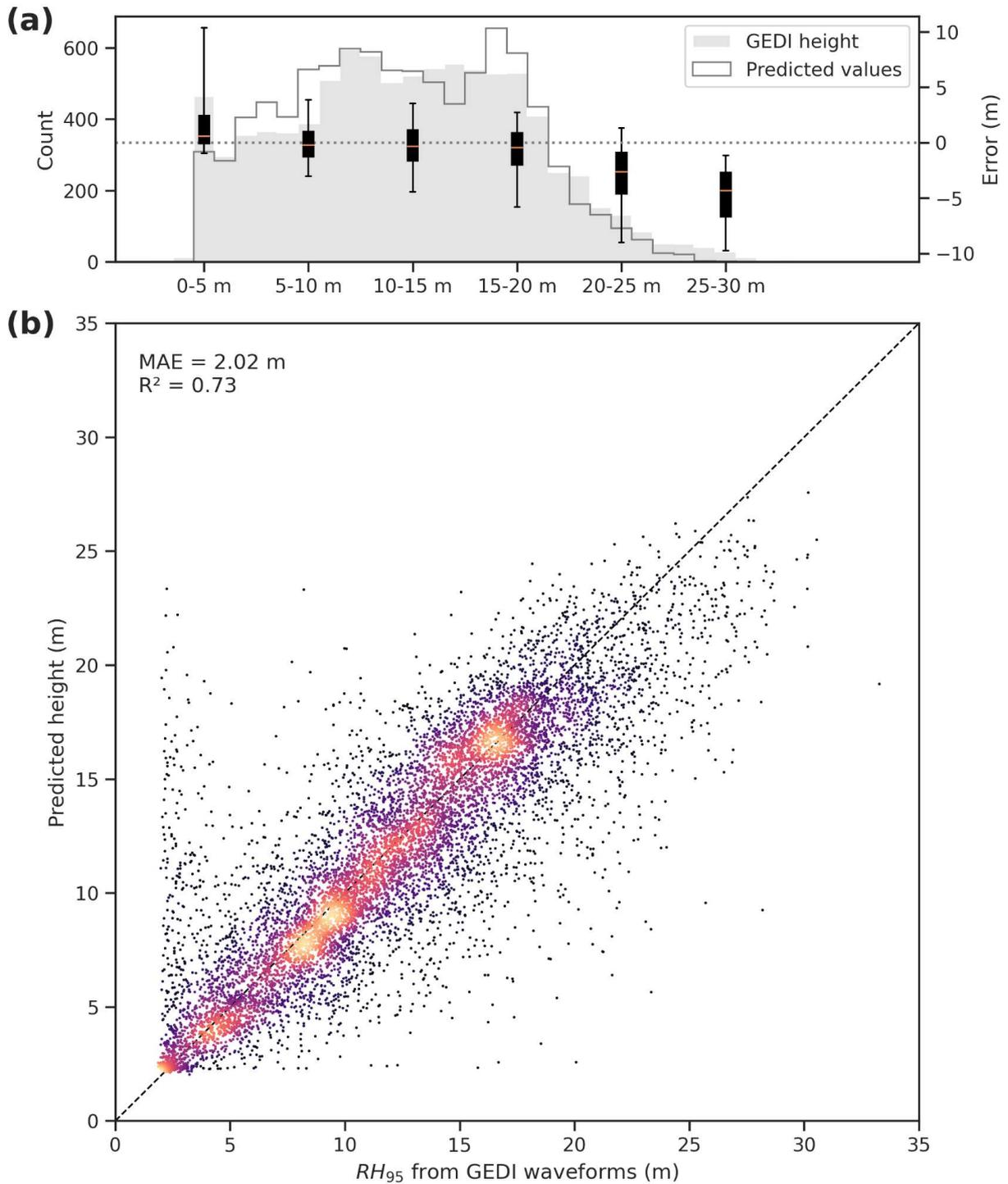

*Figure 8. Comparison of the GEDI height ($RH_{95}$) in the Test dataset and predicted values from the FCN prediction model for Scenario 1. (a) Histogram with box plots that show the differences between predicted and GEDI height per height range of 5 meters. The red lines represent median values. The upper and lower edges are the upper and lower quartiles and whiskers symbolize the 5th and 95th percentiles. (b) Density scatterplot. The dashed line corresponds to the $x = y$ axis.*

## 3.3 Evaluation with independent datasets

### 3.3.1 Forest inventory data

The comparison with the GLORIE dataset (Fig. 9a), a dense forest inventory of homogeneous stands of maritime pine in Les Landes carried out in 2015-2016 shows a good correlation with our predicted height with a $R^2$ coefficient of 0.93 (ME = 1.56 m, MAE = 2.43 m, RMSE = 2.84 m). We observe a higher prediction bias for lower heights from 0 to 10 m. The IFN dataset (Fig. 9b), composed of more diverse inventory plots (broadleaves, coniferous, distributed all over the study area and not only with homogeneous forest stands, see Fig.2), is also well correlated to our canopy height retrievals ($R^2$ = 0.71, ME = -1.18 m, MAE = 2.67 m, RMSE = 3.55 m), especially for coniferous forests ($R^2$ =0.79, RMSE = 3.09 m). Broadleaved forests show a lower correlation and poorer error metrics: $R^2$ = 0.38, RMSE = 5.74 m. Points highlighted in red in Fig. 9 indicate clear-cuts (we checked it visually with Sentinel-2 images time series) and were not considered in the calculation of the metrics.

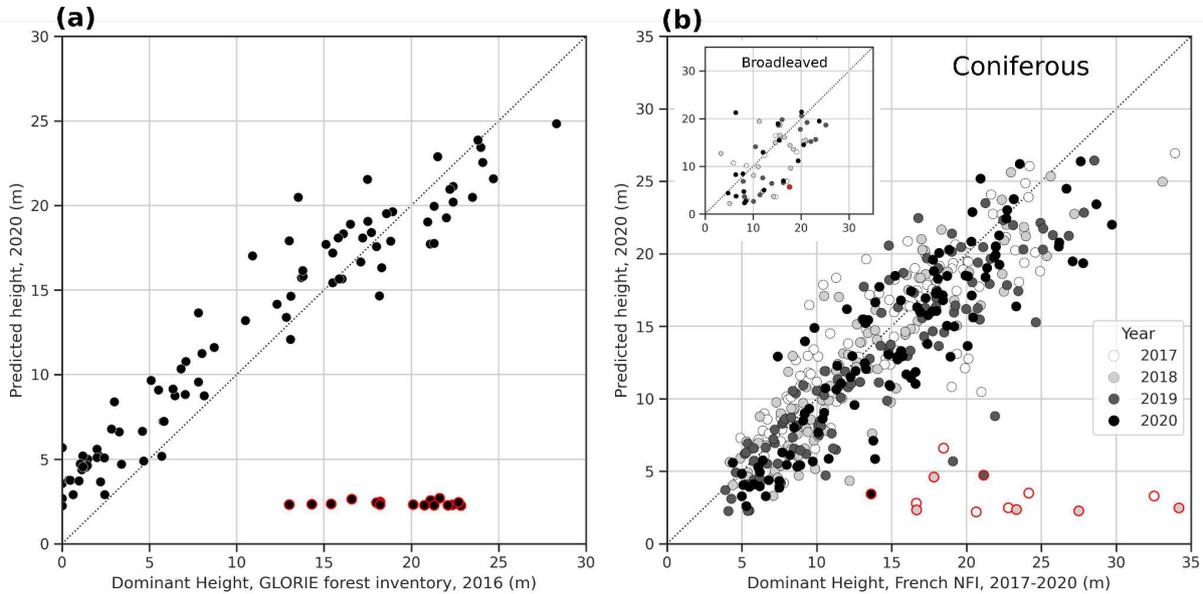

*Figure 9. Comparison between the predicted height from the FCN model (Scenario 1) for the year 2020 with the forest inventory dominant height from the GLORIE project (measurements made in 2016) (a), and with French National Forest Inventory (NFI, measurements made over 2017-2020) (b). For both graphs, the dotted line represents the x=y axis. The predicted height corresponds to the mean height of the pixels within the forest plot area. The French NFI (b) dataset is colored by year of sample and separated into broadleaved and coniferous forests. Red circles indicate plots that are not considered in the calculation of the error metrics because of forest clear-cuts between the date of the forest inventories and the date of the S1 and S2 images.*

### 3.3.2 3D reconstruction from stereo satellite acquisition

The pixel distribution in height (Fig. 10a) of our FCN model follows globally the same pattern as the 3D reconstruction canopy height model (referred to as 3D CHM in the following). The number of low height pixels between 5 m and 10 m is higher for 3D CHM while the FCN model has a higher number of pixels for all heights above 4 m, especially for heights around 16 m. The comparison at forest stand level (Fig.10b) reveals a very good correlation ($R^2$ = 0.90, MAE = 1.17 m) with no bias (ME = -0.03 m) between the 3D CHM and the FCN predictions of forest heights. Visually, the 3D reconstruction (Fig.10c) presents sharper delimitations between the different forest units compared to the FCN model (Fig. 10d). This is confirmed by the height profile (Fig. 10e) which reveals that the FCN model tends to smooth the transitions between forest units. Within forest stands, the height also seems to be averaged and more homogeneous with the FCN model compared to 3D CHM which shows more fluctuations.

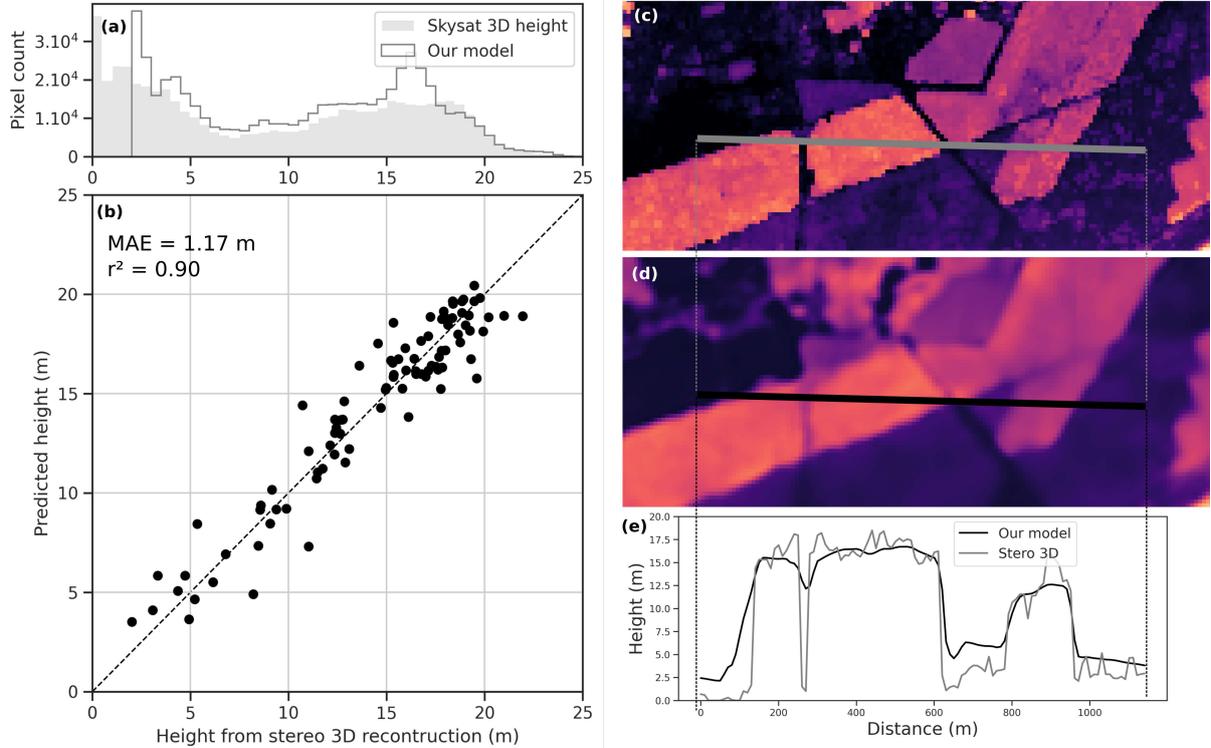

*Figure 10. Comparison of the FCN prediction model for Scenario 1 (2020) and a 3D canopy height model (3D CHM) based on Skysat imagery (2021). (a) Pixel distribution of the FCN and 3D CHM height predictions resampled at 10m with a max method. (b) Scatter plot of the comparison of the pixel median over forest stands which were labeled manually (Fig. 4). The dashed line corresponds to the x=y axis. (c) Stereo 3D reconstruction from Skysat imagery resampled at 10m (2021). (d) FCN model (2020). (e) Comparison of the FCN and 3D CHM height predictions over a 1-km height profile shown in Fig. 10c and 10d.*

## 3.4 Influence of Sentinel-1 and Sentinel-2 bands on height predictions

The results presented above are all retrieved from our best training scenario (Scenario 1: based on all S1 and S2 layers). Here, we evaluate other combinations of the S1 and S2 layers as inputs to the FCN model to understand the influence of the S1 and S2 layers in the FCN retrievals of forest heights (Fig. 11). Overall, the two first scenarios (1: All bands, 2: All Sentinel-1 bands) perform better, with a lower RMSE value when compared to any of the four evaluation datasets (*e.g.* 3.55 m and 3.76 m when compared to the French NFI while the results obtained for the other scenarios led to a RMSE > 4 m). Scenarios including only Sentinel-2 bands have lower performances (Scenarios 3, 4, and 7) but still, lead to relatively good error metrics. Similar results are observed between Scenarios 3 (all S2 bands), 4 (only 10m resolution i.e., 4 bands: RGB+NIR), and 5 (VV and VH descending from S1) except for the GLORIE dataset where Scenario 5 is better with no bias and lower SDSD. Lastly,

Scenarios 6 and 7, respectively based on one Sentinel-1 (VV descending) and one Sentinel-2 (B8: NIR) band lead to larger errors for all validation datasets (RMSE > 3 m when compared to 3D CHM while other scenarios lead to a RMSE < 2.5 m).

Overall, SB is higher when the FCN model is compared to forest inventory datasets (GLORIE and IFN). SDSD indicates how the FCN model retrieves the magnitude of the spatial fluctuations and shows higher values only for the GLORIE dataset. LCS indicates the ability of the FCN model to simulate the fluctuations across the n measurements and is the dominant term of the error for most validation datasets. For all scenarios, the correlation is better when compared to the 3D Skysat height reconstruction model.

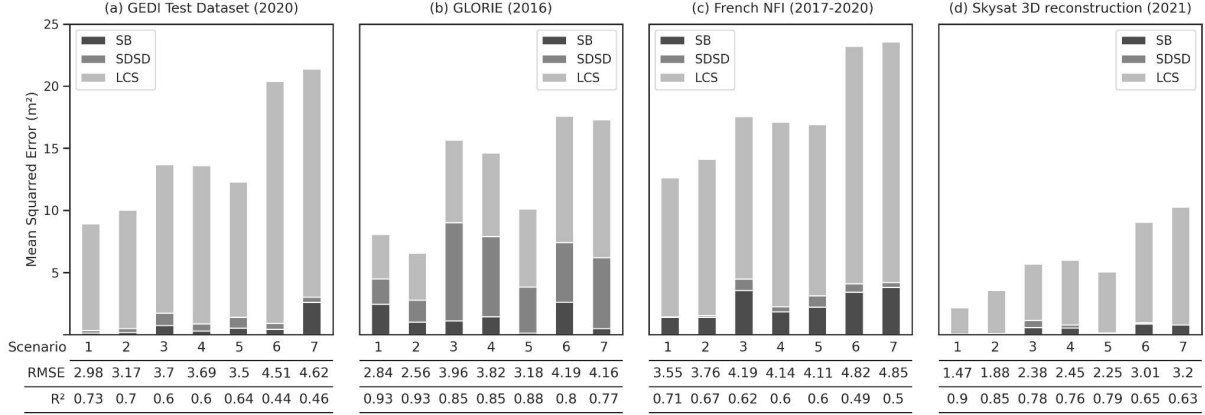

Figure 11. Comparison of the error metrics on four evaluation datasets for the 7 training scenarios. Mean squared error is decomposed into three additive terms (see 2.4.4). Squared Bias (SB) is the squared difference of the mean of both datasets. Squared Difference between Standard Deviations (SDSD) indicates how the model is able to simulate the magnitude of the fluctuations between the n measurements. The lack of correlation weighted by the standard deviations (LCS) indicates the ability of the model to simulate the fluctuations across the n measurements. RMSE and $R^2$ values are presented in the tables below the graphs. Description of the scenarios: 1 - all bands; 2 - all S1; 3 - all S2; 4 - S2 RBG + NIR; 5 - S1 VV_des + VH _des; 6-VV_des; 7-S2 NIR.

## 3.5 Comparison with other canopy height maps

In order to evaluate our results with height predictions made from models available in the literature, we compared the height predictions from the FCN model (Scenario 1, trained with all S1 and S2 layers) with three different canopy height maps: L22 (Lang et al., 2022a), P21 (Potapov et al., 2021), and M19 (Morin et al., 2019) at different locations in the Test tiles.

As an illustration, a visual comparison (Fig. 12) reveals that the FCN and L22 models predict a higher homogeneity within the forest stands while the two other models (P21 and M19) show a higher variability between adjacent pixels. Fig. 12a shows two adjacent planted forests, surrounded by a non-forested area. The GEDI footprints in this region indicate (unshown results) that the forest in the southern part of the forest plot is higher ($RH_{95}$ ~ 14 m) than that in the northern part ($RH_{95}$ ~ 10 m). This height difference is well captured by the FCN model and by L22 to a lesser extent. On the contrary, P21 shows opposite height predictions and M19 does not seem to differentiate forest heights in the southern and northern parts of the forest plot. Similarly, Fig. 12c shows an example of a more complex landscape structure with forest patches of various heights. In this case, all models are able to predict different heights for different forest stands. However, the FCN model produces more plausible outputs with clearer delimitations between forest stands of different heights. Over a broadleaved forest along a river path (Fig. 12b), L22 and P21 present higher predictions than the FCN model. M19 has not been trained on broadleaved forests and the comparison for this particular case is not relevant here.

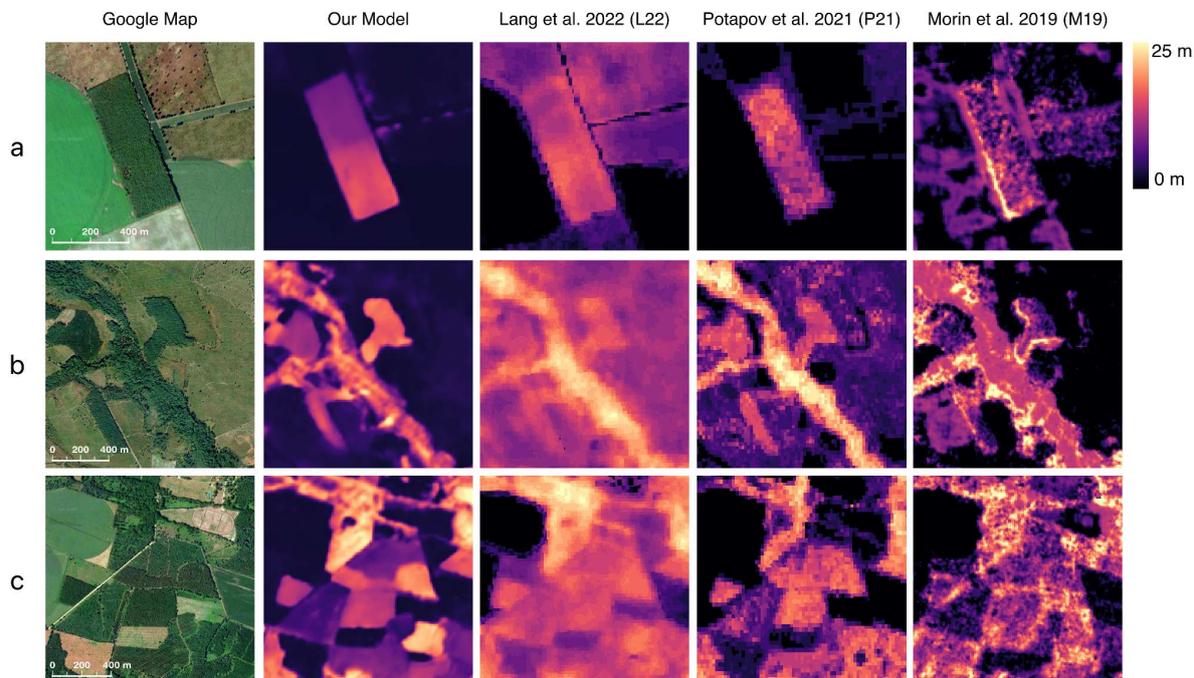

*Figure 12. Comparison of our model (FCN model) with three independent canopy height models (Lang et al., 2022a; Morin et al., 2019; Potapov et al., 2021) at three different locations in the Test tiles. (a) Two homogeneous forest stands of maritime pines with different heights. (b) Broadleaved forest along a river path. (c) Several forest stands of maritime pines with different heights.*

The comparison of the three height models (L22, P21 and M19) with the GLORIE forest inventory, the French NFI (separated into coniferous and broadleaved) and the Skysat 3D CHM (Fig. 13 and Fig. 14) reveals that the FCN model present better error metrics except for broadleaved forests. The FCN model has a RMSE of 3.09 m when compared to the French NFI coniferous stands (Fig. 13b) while L22 (RMSE = 5.01 m), P21 (RMSE = 6.57 m), and M19 (6.75 m) show higher errors. However, L22 has better performances over broadleaved forests (RMSE = 5.26 m, $R^2$ = 0.42, Fig.13c) compared to the FCN model (RMSE = 5.74, $R^2$ = 0.38). For coniferous forests, the FCN model, along with M19 shows a much lower bias (almost null for the 3D CHM) than the two global canopy height maps (P21 and L22). L22 has a high SB value (high bias) when compared to the GLORIE forest inventory and to the Skysat 3D CHM but still has a good correlation with these datasets ($R^2$ = 0.81 for GLORIE with a low LCS). This can also be visually seen in predictions in Fig. 12 and scatterplots in Fig. 14. All canopy height models have better performances on the Skysat 3D reconstruction (all RMSE values below 5 m, Fig. 13d). P19 presents a saturation effect for coniferous forests at around 15 m for coniferous forests only (Fig. 14).

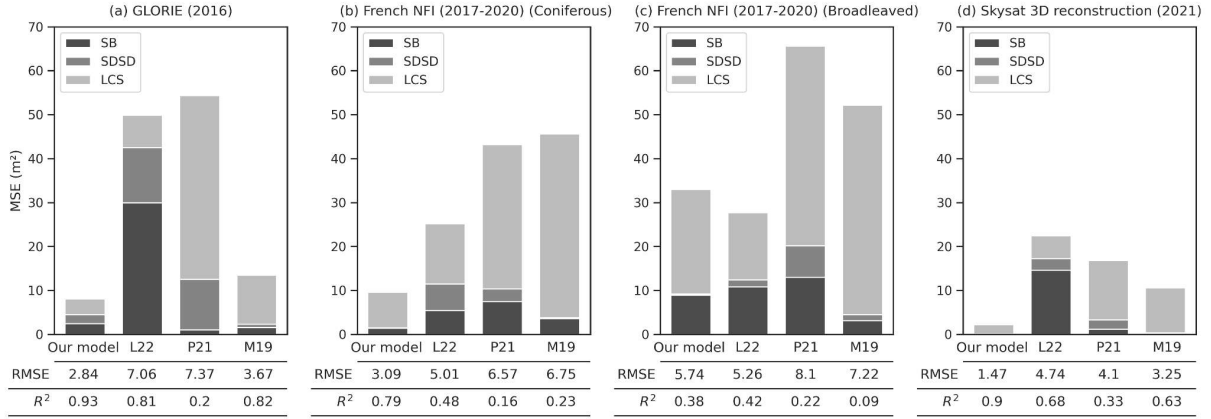

*Figure 13. Comparison of the error metrics for the different models: FCN (our model), L22 (Lang et al., 2022a), P21 (Potapov et al., 2021), and M19 (Morin et al., 2019) against data from the GLORIE forest inventory (a), the French NFI separated into coniferous (b) and broadleaved (c) forests and the stereo 3D reconstruction of Skysat imagery (d). The mean squared error (MSE) is decomposed into three additive terms (see 2.4.4). RMSE and $R^2$ values are presented in the tables below the graphs.*

L22 shows a higher correlation ($R^2 = 0.52$ for coniferous forests) than the FCN model (Fig. 15) but tends to predict higher heights (ME = 3.75 m for coniferous forests and ME = 6.0 m for broadleaved forests). The bias observed in the previous figures is also visible in Fig. 15, especially for lower heights. A lower correlation is obtained with M19 ($R^2 = 0.28$ for coniferous forests only) but with almost no bias (ME = (-0.3 m) and the predictions seem to follow the same pattern but with a high variability around the x=y axis. The comparison with broadleaved forests for M19 is not relevant as M19 has not been trained on such types of forests. P21 predicts most coniferous forests at ~ 15 m, most broadleaved forests at ~ 22 m and most other pixels at ~ 5 m.

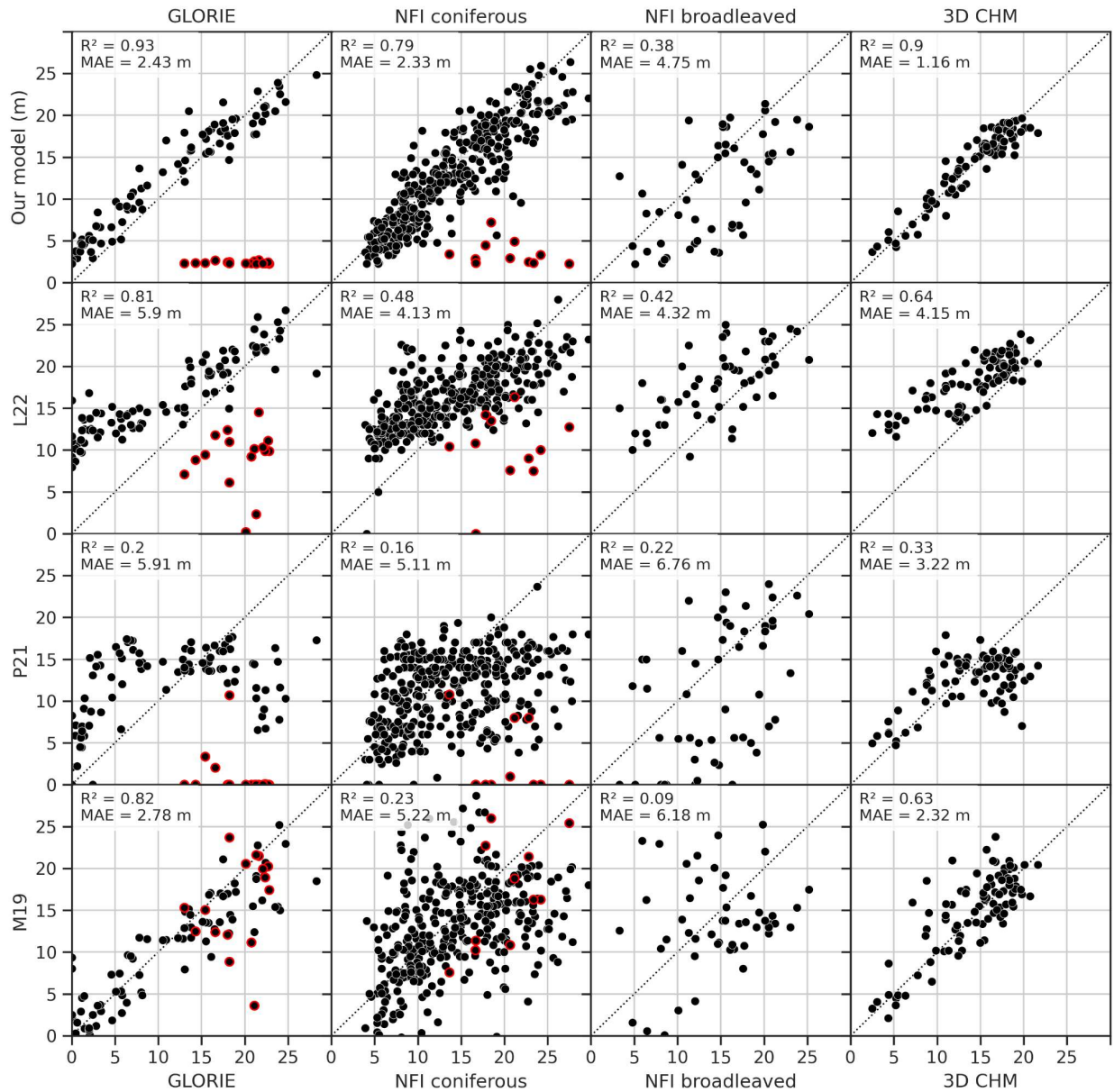

*Figure 14. Evaluation of the FCN model (Scenario 1), L22 (Lang et al., 2022a), P21 (Potapov et al., 2021), and M19 (Morin et al., 2019) against data from the GLORIE forest inventory, the French NFI (separated into broadleaved and coniferous) and the 3D height reconstruction from Skysat imagery. Points circled in red indicate forest clearcuts between the date of inventory and 2020. The latter data are not considered in the calculation of $R^2$ and MAE.*

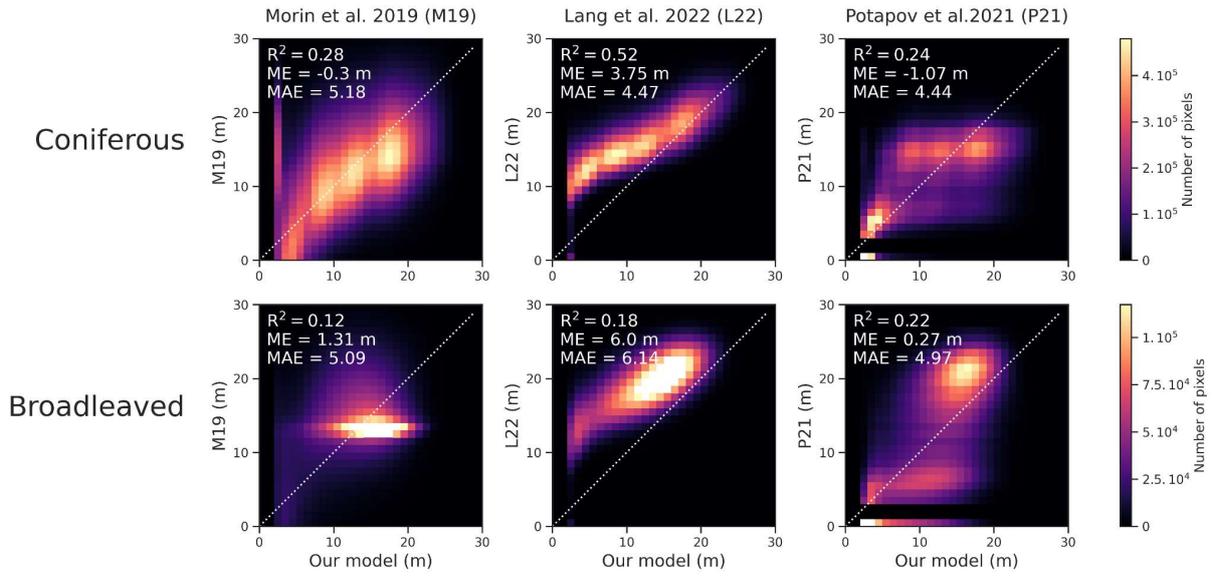

*Figure 15. Pixel-wise comparison of the FCN model with three independent canopy height models: P22, L21 and M19 (Lang et al., 2022a; Morin et al., 2019; Potapov et al., 2021). Forest pixels were categorized into coniferous or broadleaf forest with the Copernicus forest type map. The white line represents the x=y axis.*

# 4 Discussion

- Forest border errors

The visual analysis of the FCN height predictions shows higher errors at the forest borders (Fig. 6). These errors are likely related to the GEDI uncertainty of ~ 10 m for the ground location. Some GEDI footprints are located within a forest, close to the border, but the waveform corresponds to the reflection on bare soil outside the forest. This effect may also explain the high FCN height predictions when the $RH_{95}$ values are close to zero in Fig. 8. However, the opposite phenomenon (low predictions for high $RH_{95}$ values) is observed less often, which is likely due to the fact that the GEDI Test dataset we used only contains footprints that are located within forests. As the GEDI's footprint diameter is 25 m, small gaps in the canopy that are invisible in the S1 and S2 images cannot be reflected in GEDI's $RH_{95}$ and cannot explain these high prediction errors. To a lesser extent, clearcuts occuring between the date of the GEDI acquisition and the date of the S1 and S2 images could be responsible for some of these errors.

- Smoothness

On the border of landscape units, a smoothing effect seems to occur in the predictions. This can be particularly observed in the comparison with the 3D reconstruction from stereo Skysat images (Fig. 10c,d,e). Contrary to a per pixel prediction, as performed in P21 for instance, each pixel of the FCN canopy height map is the result from multiple convolutions involving the surrounding pixels (See 2.4.1 for U-Net structure). Pixels close to forest borders have neighbor pixels from forest and bare soil which could lead to this average height prediction, creating smoothness between landscapes units. Additionally, the 10 m x 10 m S1 and S2 pixels on forest borders contain also average information from both forest and non-forest areas and are also "smoothed" which can explain the smoothing effect in our predictions (Fig. 16). Finally, it is likely the FCN model tends to predict average values at forest borders to avoid making high errors during the training process because of the GEDI location uncertainty.

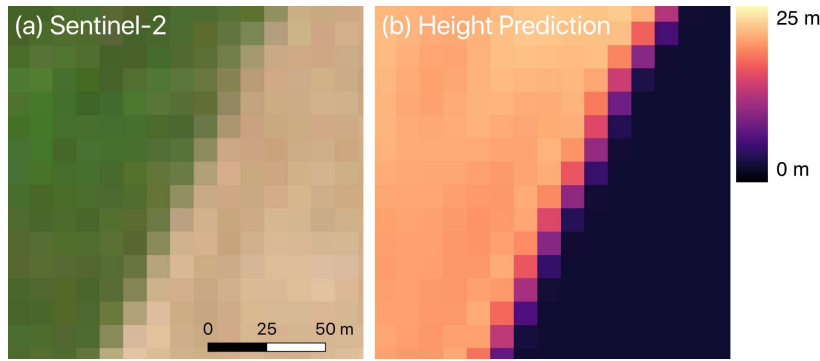

*Figure 16. (a) Sentinel-2 composite image used for prediction. The transition between the forest and non-forest area is not sharp. Some 10m x 10m pixels contain average information from both landscape units. (b) Height prediction from the FCN model (Scenario 1). The transition between the two landscape units is "smoothed".*

- Bare soil

Due to $RH_{95}$ properties, bare soil is retrieved with a height of ~2.25 m (See 2.2.1) and makes the FCN canopy height model not suitable below this height. The height profile (Fig.10.e) used to compare the 3D reconstruction with the FCN model highlights this phenomenon where we can see that low non-forest heights are overestimated. The overestimation of lower heights visible in Fig. 9a for the GLORIE dataset is mainly related to this $RH_{95}$ effect and explains the higher SDSD values in Fig. 11b. Indeed, this dataset contains several forest plots with tree height measurements close to zero (no forest in some cases or newly planted forest in other cases). A tree growth effect, related to the difference of four years between the inventory (2016) and the prediction (2020), accentuates this phenomenon.

- Underestimation at high forest heights

Our model underestimates heights above 20 m (ME = -2.5 m for trees between 20 and 25 m, Fig. 8a). This phenomenon is common in most studies that try to predict forest height from machine learning algorithms but do not always occur at the same height (Lang et al., 2022a, 2019; Morin et al., 2019; Potapov et al., 2021). Rather than a saturation effect related to the information contained in S1 and S2, it is more likely due to an imbalanced distribution of reference height labels. Indeed, in the Test dataset (only forests), GEDI $RH_{95}$ values above 20 m account only for 8.9% of the total number of footprints and 1.5% for $RH_{95} > 25$ m. Several techniques such as a weighted cost function can be used to reduce this effect and will be investigated in future developments of the FCN model.

- Better performance over coniferous forests

Thanks to the French NFI data, we were able to evaluate separately the performance of the FCN model over broadleaved and coniferous forests. The error metrics are much better for coniferous forests for several reasons. As well as for higher heights, the number of reference height data for broadleaved forests was much lower in our data sets than for maritime pine forests (only 10% of the forested areas in the Landes forest is not covered by maritime pines). Additionally, these broadleaved forests have a more complex structure, with a higher spatial heterogeneity. Within one 25 m GEDI footprint, several tree species, with different heights and shapes are summed up into one $RH_{95}$ value. Therefore, an error on the GEDI footprint location has a higher impact on model training for this type of forest. This complexity also makes the on-site measurement of tree height more difficult which results in less precise validation data for broadleaved forests.

- Influence of the S1 and S2 layers

The analysis of the seven training scenarios (Fig. 11) reveals that the more S1 & S2 layers are available as inputs, the better the prediction is. However, training the model only with a subset of the 14 layers proposed as inputs (10 for S2, 4 for S1) still leads to very satisfying error metrics. The S1 bands seem to be the most interesting predictors in our case (Scenario 2, RMSE = 3.76 m on French NFI), and adding 10 S2 bands (Scenario 1) decreases the RMSE to 3.55 m only. But a model trained only with S2 (Scenario 3) is still able to carry out good predictions (RMSE = 4.19 m on French NFI). Scenario 4 highlights that only the S2 spectral bands with a 10 m resolution are necessary to obtain such results which can be related to the higher resolution of these bands but also to the contained information (RGB + NIR). A combination of two S1 layers (Scenario 5) leads to much better results than a single layer (Scenario 6). In other results that were not presented here, we found that any combination of two S1 bands led to better results than the best S1 band alone (VH descending). Hence, increasing the number of input bands improves the predictive power of the model. Even though they might be very specific to the very particular forest structure of the Landes forest, these results suggest that, for highly cloudy regions where cloud-free S2 images are rare, a training based only on S1 could be sufficient to retrieve forest height.

- Comparison with other canopy height models

For most evaluation metrics, the FCN model shows improved performance compared to available canopy height maps for the Landes forest. Better delimitation between landscape units, lower bias, better correlation, and better RMSE are obtained on all validation datasets except for the broadleaved part of the NFI data set where L22 is better. These results underline the importance of the scale at which the training is performed. P21 and L22 are the outputs from global models, trained with optical imagery (Landsat-8 for P21, S2 for L22) and GEDI metrics as reference height. Even though the number of GEDI data used specifically for the Landes forest by P21 and L22 had the same order of magnitude as our Train dataset, they only represent a very small fraction of the global GEDI dataset. The P21 and P22 models were trained to optimize height prediction globally and can make predictions on more diverse forest types but they are less specific to the Landes forest. For instance, L22 performs slightly better on broadleaved forests which is most likely related to the higher number of reference height data for this type of forest available in the global dataset used for training. It is likely the bias observed for lower heights comes too from this global training process.

The comparison with M19 highlights the importance of representativity and quantity of reference height data. In this case, the reference data used for training M19 (GLORIE dataset only) is only representative of homogeneous areas within maritime pine plantation stands. Therefore, the M19 model did not learn to deal with different forest surfaces such as borders, gaps or broadleaved forests. This model shows good performances on the 3D Skysat reconstruction which is close to the area where it was trained and where we aggregated the height values at the forest stand level. Lower performances are observed on the French NFI, probably because some of the forest plots are close to forest borders and not located close to the training area. Additionally, the GLORIE forest inventory (see 2.3.1) contains only 99 height samples while we used 175511 valid GEDI waveforms to train our FCN model.

Deep learning techniques applied to images (CNN) have the advantage of being "spatially aware" while other classical machine learning methods (Random Forest for P21, Support Vector Machine for M19) are not. In P21 and M19, forest texture metrics were added to the prediction variables in order to give this spatial awareness to the model but it is limited compared to the high number of convolution filters within the U-Net model that we used. In a newer study, Morin et al. (2022) have integrated GEDI features into their forest parameters retrieval method. Another comparison with these new methods would be interesting to understand whether it is deep learning (compared to simpler machine learning algorithms) or the number of training data (from GEDI) that makes our current model better than M19.

- On the use of GEDI

The GEDI mission provides an unprecedented database of LiDAR waveforms with high precision and has a high potential for forest parameter estimation. However, some caveats remain when using this type of data as ground truth for deep learning models. Indeed, the ability of GEDI to retrieve canopy height properly in more complex forest structures is still uncertain. In denser forests, some GEDI laser beams could not penetrate deep enough in the trees to reach the ground thus leading to a height underestimation. Moreover, the uncertainty associated with the GEDI footprint location could potentially lead to large errors. Here, we rasterized the GEDI footprint in a 10 m x 10 m pixel that corresponds to the center of this footprint. However, the height information contained in this GEDI footprint encompasses a 25 m diameter circle. Additionally, the 10 m uncertainty on the footprint location extends the area where the height information is actually captured by GEDI to a 45 m diameter circle around the supposed center of the footprint. Hence the height information could potentially come from a point that is the third neighbor of the actual pixel that was rasterized. These errors are randomly distributed and, thanks to the large number of GEDI samples, it only creates noise in the reference height data used for training. So, it seems that these errors can be well handled by the FCN model. But in other more complex regions, this noise can be higher and the FCN model could have difficulties to carry out correct height predictions. To evaluate the impact of this uncertainty, we retrained our model with the previous GEDI release (GEDI v001). In this previous version, the standard deviation on the GEDI footprint location was 20 m (GEDI v001) and it was improved to 10 m in the GEDI v002 version. The MAE on the Test dataset improved from 2.43 m to 2.02 m when using the more recent version of GEDI. FCN model outputs show clearer patterns and transitions between forested and non-forested areas for v002 *vs* v001 (See Appendix 4). The FCN model based on GEDIv001 produces blurred transitions between landscape units, thus trying to avoid high losses related to location errors. A further improvement in GEDI location has the potential to increase even more the accuracy of the FCN height maps we produced in this study.

- Influence of the number of training samples

The GEDI footprints are unevenly distributed globally because of the ISS trajectory. Therefore, in some other regions, fewer waveforms would be available to apply the same methodology. In order to assess how this can affect the reproducibility of the proposed method, we trained three additional models like the one from scenario 1 (all S1 and S2 bands) by randomly keeping only 10%, 1% and 0.1% of the footprints from the training dataset. As shown in Table 1, a model trained with only 10% of the original train dataset still leads to good results (MAE = 2.43 m on the Test dataset) and a canopy height map that looked very similar to the original one by visual inspection. However, when the number of training samples drops to 1% of the original sample size, the output map looks a little noisier with a lower MAE (3.02 m).

| Size of the training sample (% of the original training sample size) | 100% | 10% | 1% | 0,1% |
|---|---|---|---|---|
| MAE on Test dataset | 2.02 m | 2.43 m | 3.02 m | 3.94 m |
| 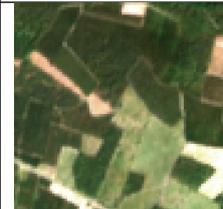 | 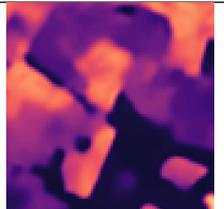 | 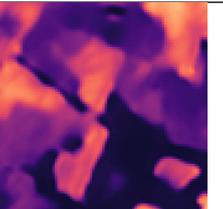 | 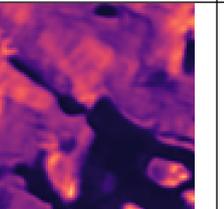 | |

*Table 1. Evolution of the MAE on the Test dataset for 4 different FCN models trained with 100% (131.633 footprints), 10%, 1% and 0.1% of the original Train dataset size. All the models correspond to scenario 1 (All S1 and S2 bands). The second line shows an example of the height prediction, with the corresponding S2 image in the first column.*

These results stress the importance of having large training datasets, especially for deep learning algorithms like our FCN model. However, the model still has good performances with only 10% of the original Train dataset size which makes this methodology robust and suitable for being applied in other regions with a lower amount of GEDI data.

# 5 Concluding remarks

This study highlights the potential of deep learning models to map forest height continuously at high resolution with sparse references, *e.g.* GEDI LiDAR data, and continuous images from Sentinel-1 SAR and Sentinel-2 multispectral imager. Additionally, it confirms the potential of GEDI data to produce a good estimation of forest height, especially when integrated into a deep learning prediction model that reduces the uncertainty related to the measurement of one LiDAR footprint. Our FCN model is able to retrieve forest height at 10 m resolution in a French coniferous plantation with a relatively good accuracy (MAE = 2.02 m) for the GEDI Test dataset and a high correlation with independent height measurement sources. These results remain relatively good when using only one satellite source for model training which is particularly interesting in order to extend this study to other regions of the globe with a higher cloud cover or to use other satellite sources like Planet that have only four spectral bands but a higher spatial resolution. The map we produced showed improved results in comparison to other existing canopy height models over this region. The Landes forest is mainly composed of even-aged forest parcels of maritime pines. This very specific situation has probably been "learned" by the deep learning model which tends to smooth the height within one forest stand and its performance on other types of forest structures is probably different. However, the method we developed still appears to be promising to retrieve canopy height on other forest structures such as broadleaved forests. For instance, we applied this methodology to another French "sylvoecoregion" called Sologne, mostly composed of broadleaved forest and obtained a similar accuracy with external validation from NFI plots (MAE = 2.62 m, see Appendix 6), Additionally, the area of interest is mostly flat. Sentinel-1 images are very sensitive to steep terrain and GEDI waveforms are also affected by higher slopes that tend to overestimate tree heights (Kutchartt et al., 2022). Further studies in more mountainous regions, by including a digital surface model in the deep learning process, could be potentially interesting to address these issues. Considering the availability of the Sentinel-1 and -2 observations, the canopy height map we retrieved could be used to monitor tree height with a yearly time frequency. Moreover, considering the strong relationship between canopy height and canopy biomass (Saatchi et al., 2011), the canopy height map could be used in a subsequent step to monitor forest biomass at the forest stand level and with a good temporal repetition (at least yearly), thus following the guidance of the Global Forest Observation Initiative (GFOI) to integrate earth observation data into national forest monitoring systems.

# 6 Supplementary material

The canopy height map for 2020 of Les Landes is accessible on this Google Earth Engine online app: https://martinschwartz0.users.earthengine.app/view/chm and can be used and downloaded from this plateform with the following asset ID: "projects/ee-martinschwartz0/assets/CHM_Landes_MSchwartz"

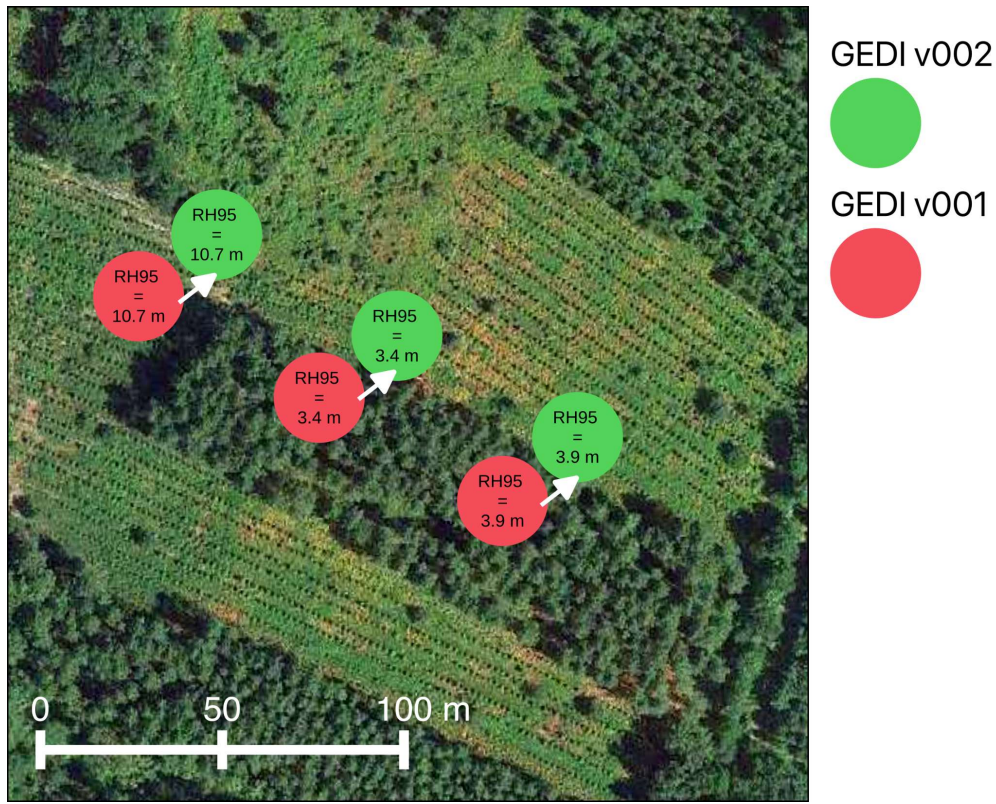

*Appendix 1: Examples of GEDI v001 and GEDI v002 footprints (25 m of diameter) on a coniferous forest parcel in the Landes forest. The white arrows represent the shift in position operated in v002 which is more consistent with the landscape observed here.*

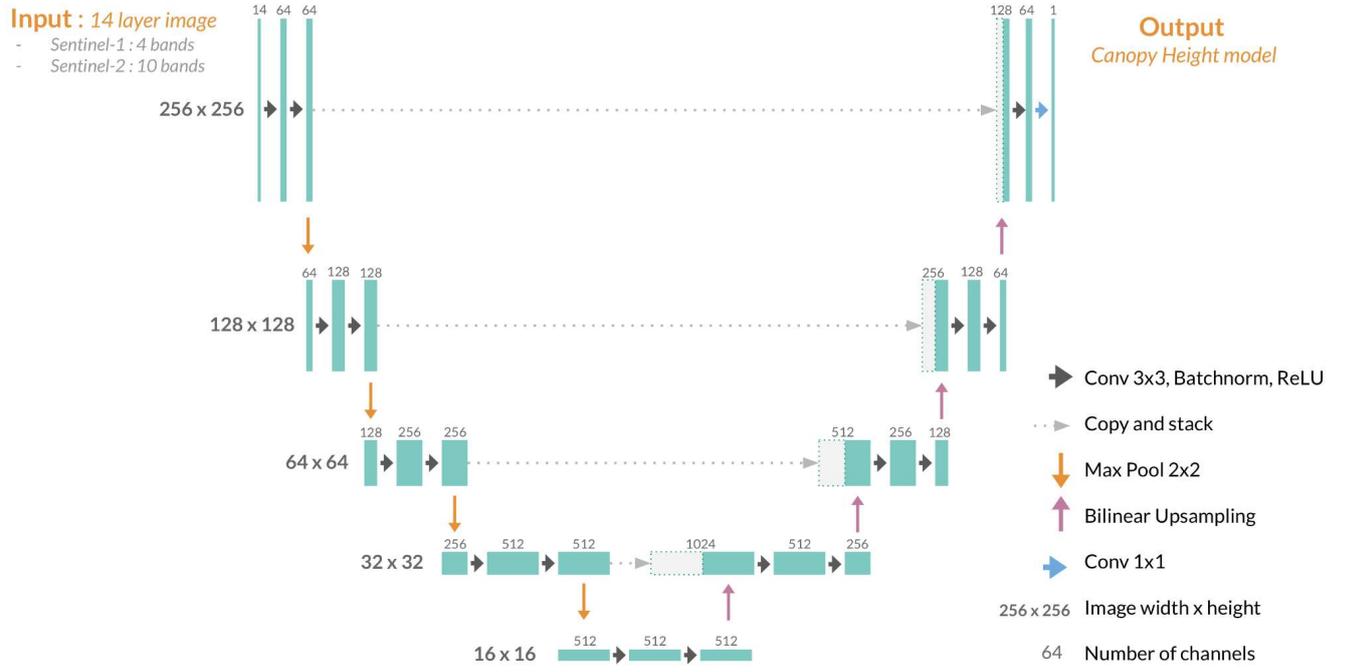

*Appendix 2. Structure of the U-Net model used.*

$$MAE = \frac{\sum_N |h_{pred} - h_{val}|}{N} \qquad RMSE = \sqrt{\frac{\sum_N (h_{pred} - h_{val})^2}{N}}$$

$$ME = \frac{\sum_N h_{pred} - h_{val}}{N} \qquad R^2 = \left(\frac{cov(h_{pred}, h_{val})}{\sigma_{h_{pred}} \times \sigma_{h_{val}}}\right)^2$$

$$MSD = RMSE^2 = SB + SDSD + LCS$$

$$SD_s = \sqrt{\frac{1}{n}\sum_{i=1}^{n}(x_i - \bar{x})^2} \qquad SD_m = \sqrt{\frac{1}{n}\sum_{i=1}^{n}(y_i - \bar{y})^2}$$

$$SB = (\bar{x} - \bar{y})^2 \qquad SDSD = (SD_s - SD_m)^2 \qquad LCS = 2SD_s SD_m (1 - r)$$

*Appendix 3. Detailed equations of the error metrics presented in this study. hpred is the predicted height, hval is the height from the validation dataset, N is the number of height samples, cov the covariance function, and σ the standard deviation of the datasets. MSD is decomposed into three additive terms as defined in (Kobayashi and Salam, 2000). $SD_s$ is the standard deviation of the predicted data while $SD_m$ is the standard deviation of the reference or evaluation data.*

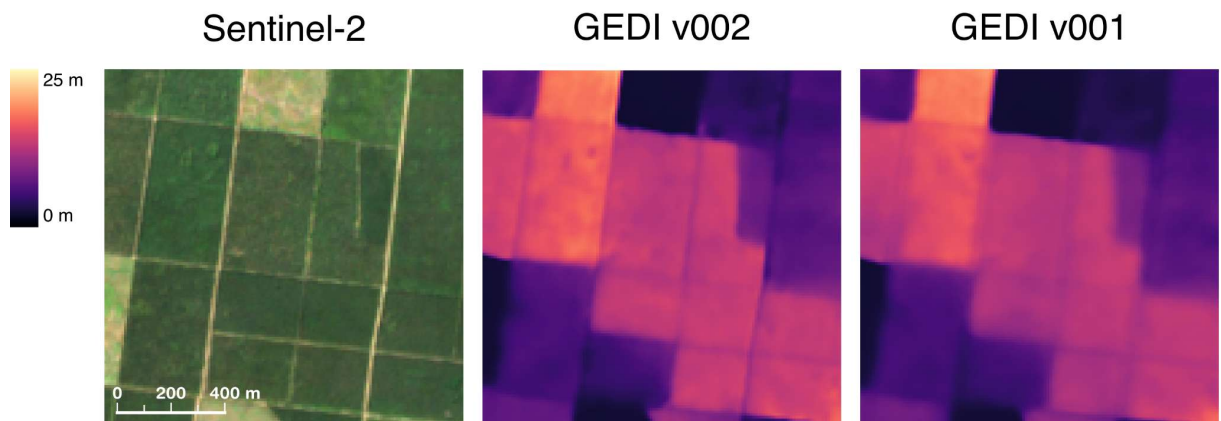

Appendix 4. Example of model output trained with GEDI v001 (20 m precision on footprint location) and GEDI v002 (10 m precision on footprint location). The model trained with more accurate GEDI data has cleaner results, with less smoothing at forest transitions.

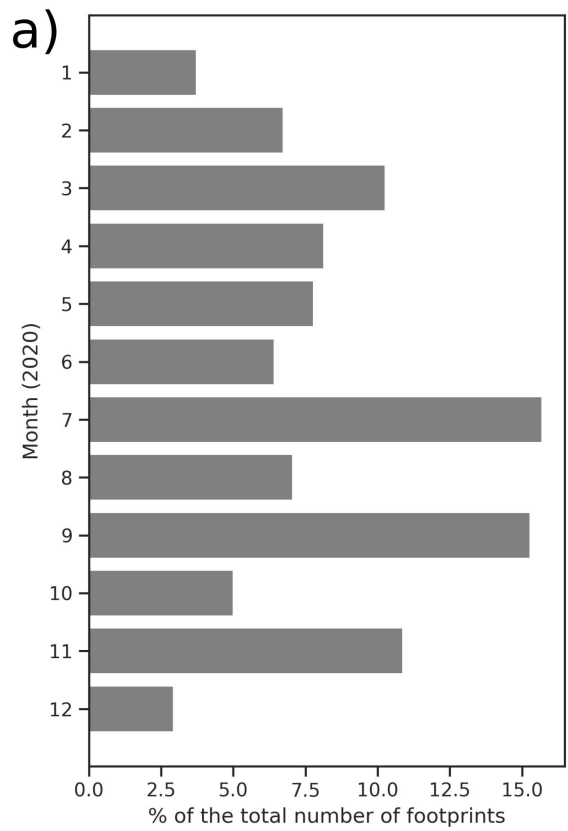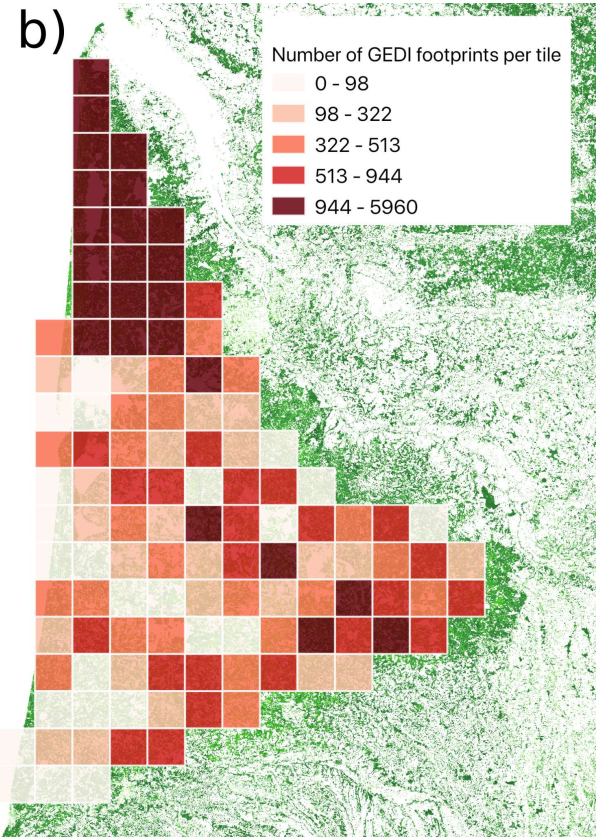

*Appendix 5. (a) Monthly distribution of the GEDI footprints after filtering (Section 2.2.1). (b) Spatial distribution of the GEDI footprints after filtering. Each color accounts for 20% of the tiles. More GEDI footprints are located in the northern tiles due to the ISS trajectory.*

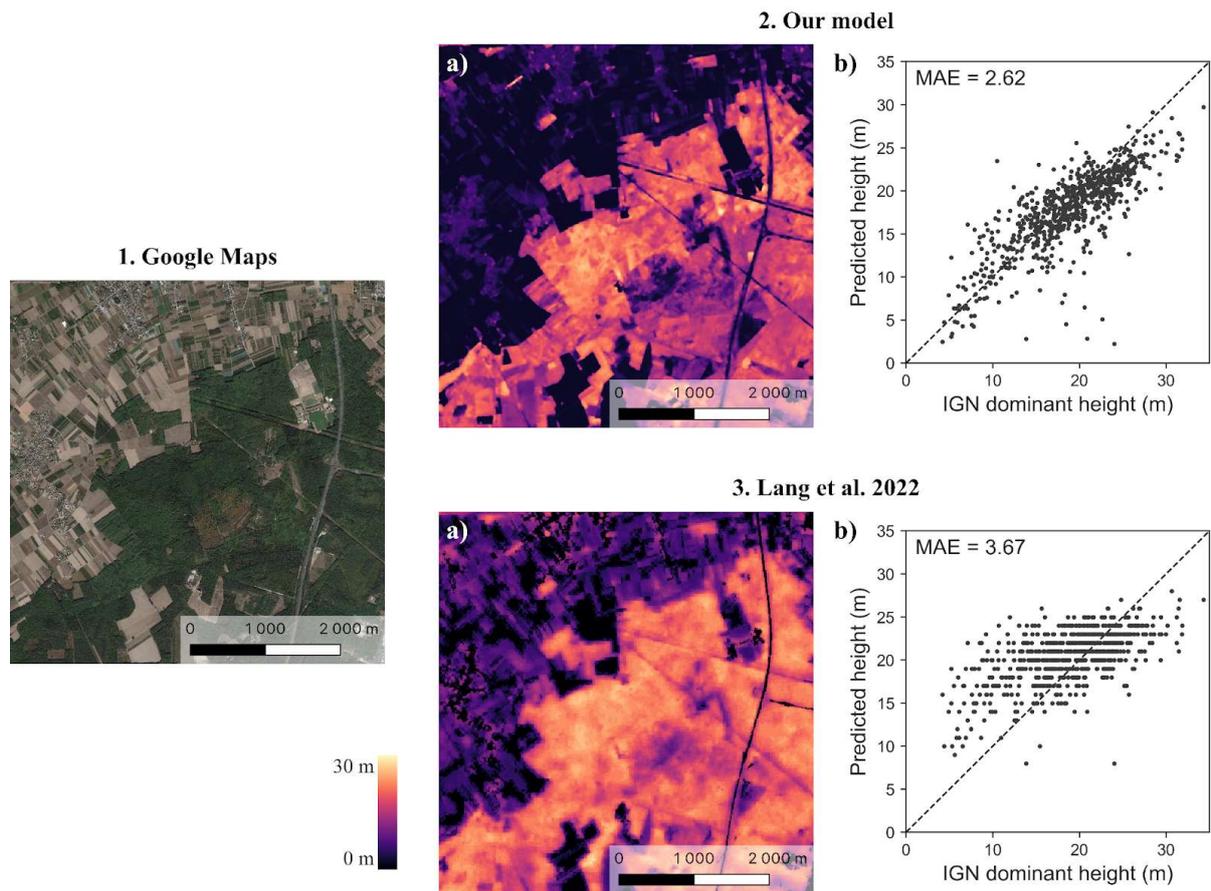

*Appendix 6. Comparison between the height from another FCN model trained in Sologne (France), the height from the French forest inventory plots (IGN) in Sologne where 75% of the plots are in broadleaved forests, and the height from the global study of L22 (Lang et al., 2022). (1) Google map image. (2) a.Canopy height map from our FCN model at 10 m resolution. b.Comparison with the French NFI plots (MAE = 2.62 m) (3) a.Canopy height map from L22 at 10 m resolution. b.Comparison with the French NFI plots (MAE = 3.67 m).*

## Appendix 7: Scale mismatch between GEDI and Sentinel data

GEDI footprints corresponds to 25 m diameter circles while S1/S2 data have a pixel size of 10 m. This scale mismatch can have an influence on our results and it can be argued that the true resolution of our map is closer to 25 m than 10 m. To test this hypothesis, we designed another FCN model (the "modified model" in the following) where a 2x2 convolution with stride 2 has replaced the last layer. Because of this last layer, the output of the modified model is at a 20 m resolution. We trained this model with GEDI data rasterized on a 20 m grid, hence reducing the scaling issue (See Appendix 7.1).

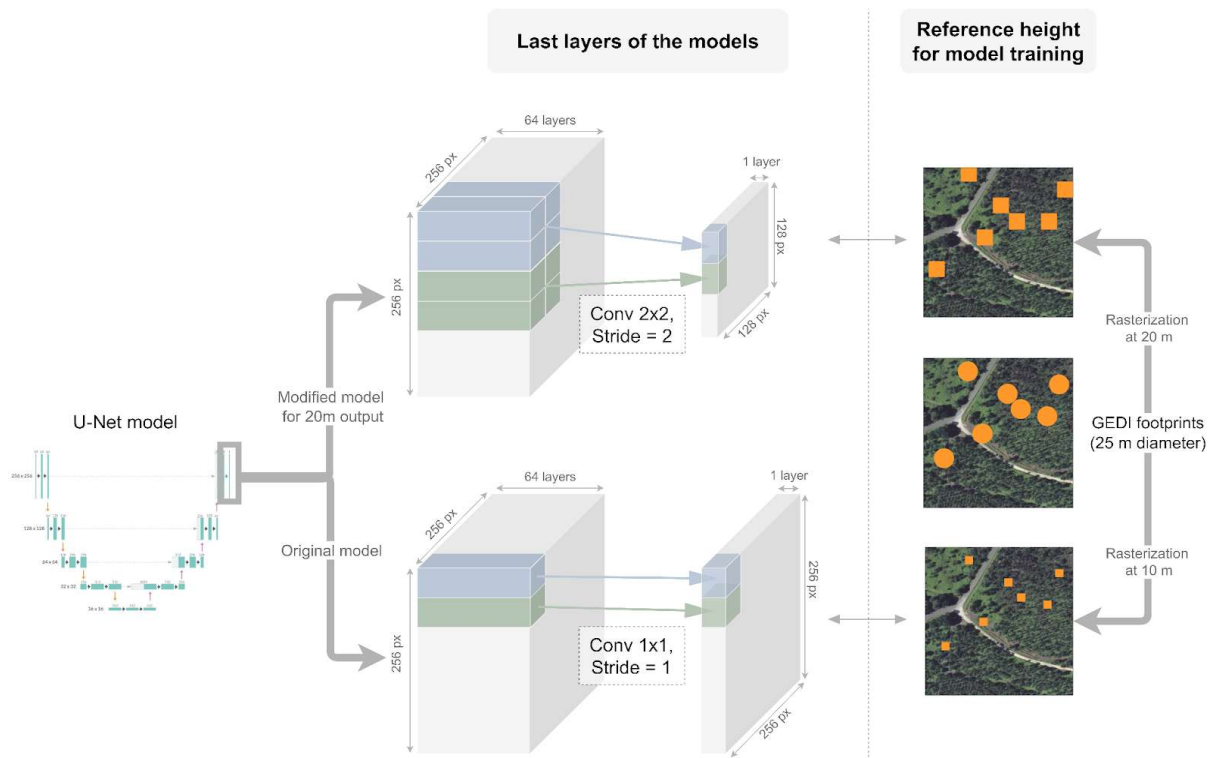

*Appendix 7.1: Details of the two last layers of the original and the modified models. In the modified model, the last 1x1 convolution is replaced by a 2x2 convolution with a stride value of 2, hence creating a 20 m resolution output that can be compared to GEDI footprints rasterized at 20 m to train the model. The reference height data used for the 20 m resolution model are much closer to the footprint size of GEDI than the one used for the original model, hence addressing the scaling issue.*

To proceed to a fair comparison with the output of our model that has a native pixel size of 10 m, we applied a bilinear upsampling to the output of the modified model, thus creating a map with 10 m pixels (Appendix 7.2).

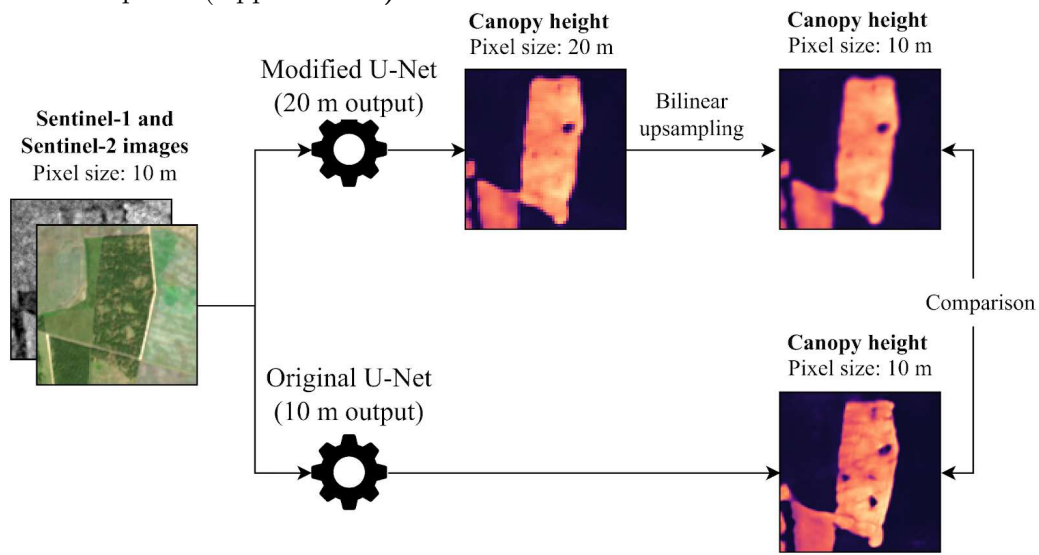

*Appendix 7.2: Workflow proposed to compare the model trained with GEDI footprints rasterized at 20 m and the model that used GEDI footprints rasterized at 10 m. The output map at 20 m is resampled at 10 m in order to create a 10 m map for a fair comparison to our original product.*

Compared to the GEDI test dataset, our original model (MAE = 2.02 m) performs slightly better than the modified model (MAE = 2.19 m). In addition, we can visually see that the original 10 m model is more able to retrieve heterogeneous canopy cover. As an illustration, in the following examples (Appendix 7.3), the red squares indicate holes in the canopy that the modified model at 20 m could not capture.

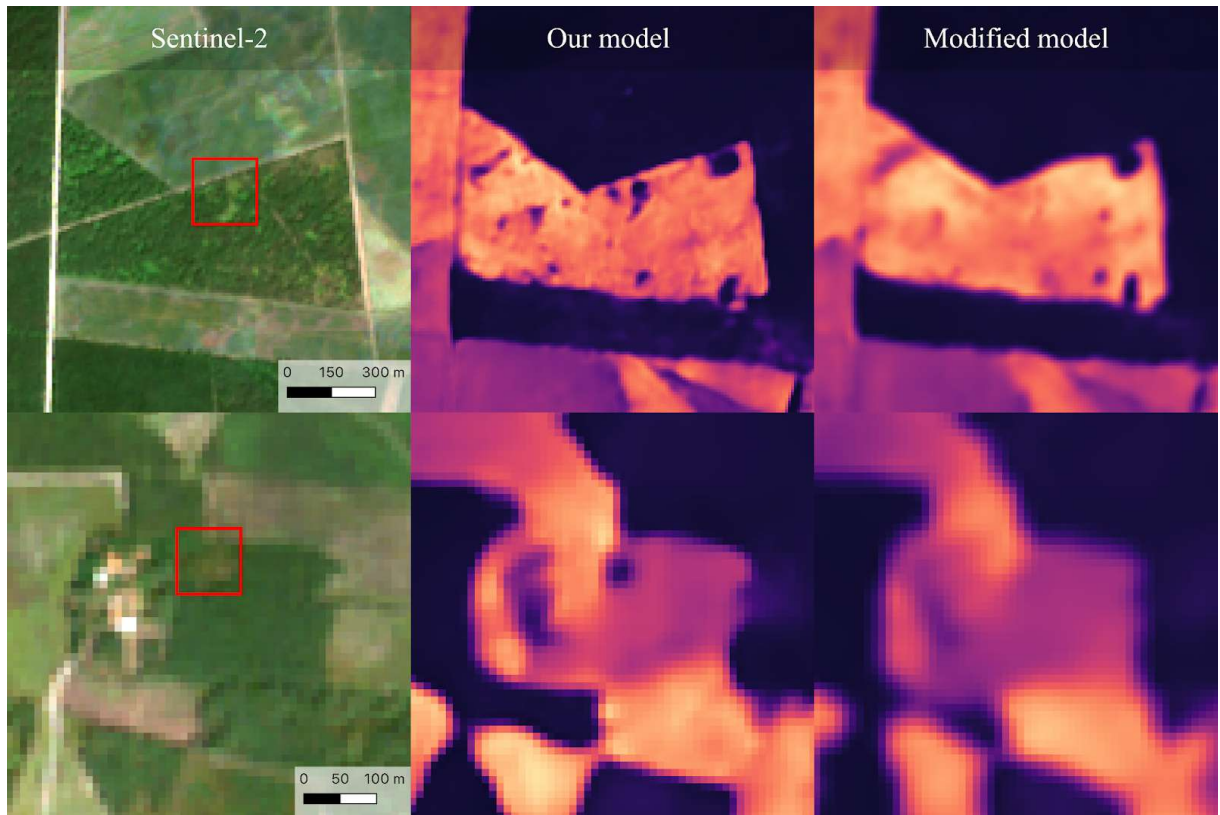

*Appendix 7.3: Visual comparison between Sentinel-2 images, our model trained at 10 m resolution and the modified model trained at 20 m resolution here upsampled at 10 m for a fair comparison. Complex forest structures in the canopy indicated by red squares are not well retrieved in the modified model outputs, while our model can retrieve them. Our model has a MAE of 2.02 m for the Test dataset while the modified model has a MAE of 2.19 m.*

Based on these elements, we are quite confident to state that the label noise created by the scaling issue is really well handled by the deep learning model. Heterogeneous canopy height is well retrieved by the model, and the use of GEDI at a 10 m scale yields better results than at a 20 m scale, though it is closer to the 25 m circular footprint.